\pdfoutput=1

\documentclass[11pt]{article}

\usepackage{ACL2023}

\usepackage{times}
\usepackage{latexsym}

\usepackage[T1]{fontenc}

\usepackage[utf8]{inputenc}

\usepackage{microtype}

\usepackage{inconsolata}

\usepackage{graphicx}
\usepackage{tabularx}
\usepackage{adjustbox}
\usepackage{booktabs}
\usepackage{multirow}
\usepackage{amsmath}
\usepackage{cuted}
\usepackage{comment}
\usepackage{lipsum}
\usepackage{colortbl}
\usepackage{xcolor}
\usepackage{enumitem}

\title{Exploring the Impact of Layer Normalization for \\ Zero-shot Neural Machine Translation}

\author{
Zhuoyuan Mao $^{1}$ \hspace{1em}
Raj Dabre $^2$ \hspace{1em} 
Qianying Liu $^1$ \hspace{1em} \\
{\bf Haiyue Song $^1$} \hspace{1em}
{\bf Chenhui Chu $^1$} \hspace{1em}
{\bf Sadao Kurohashi $^{1, 3}$}\\
$^1$ Kyoto University, Japan \hspace{1em}
$^2$ NICT, Japan \hspace{1em} 
$^3$ NII, Japan \\
\texttt{\{zhuoyuanmao, ying, song, chu, kuro\}@nlp.ist.i.kyoto-u.ac.jp} \\
\texttt{raj.dabre@nict.go.jp}
}

\begin{document}

\maketitle

\begin{abstract}
This paper studies the impact of layer normalization (LayerNorm) on zero-shot translation (ZST). Recent efforts for ZST often utilize the Transformer architecture as the backbone, with LayerNorm at the input of layers (PreNorm) set as the default. However,~\citet{DBLP:conf/nips/Xu0ZZL19} has revealed that PreNorm carries the risk of overfitting the training data. Based on this, we hypothesize that PreNorm may overfit supervised directions and thus have low generalizability for ZST. Through experiments on OPUS, IWSLT, and Europarl datasets for $54$ ZST directions, we demonstrate that the original Transformer setting of LayerNorm after residual connections (PostNorm) consistently outperforms PreNorm by up to $12.3$ BLEU points. We then study the performance disparities by analyzing the differences in off-target rates and structural variations between PreNorm and PostNorm. This study highlights the need for careful consideration of the LayerNorm setting for ZST. 


\end{abstract}

\section{Introduction}
\begin{figure}
    \centering
    \includegraphics[width=\linewidth]{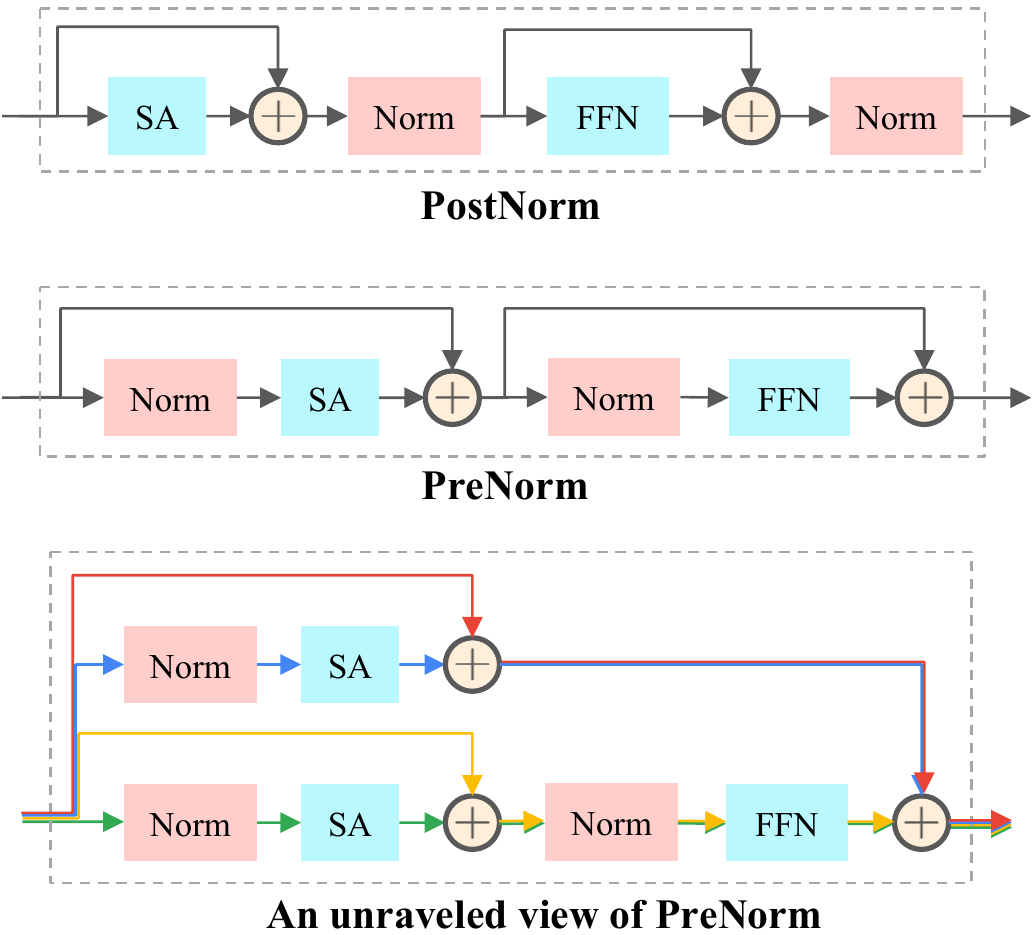}
    \caption{\textbf{PostNorm, PreNorm, and an unraveled view of PreNorm in a Transformer encoder layer}. ``Norm,'' ``SA,'' and ``FFN'' denote LayerNorm, self-attention, and feed-forward network. $\oplus$ is residual connection. Paths with different colors in the unraveled view of PreNorm indicate respective sub-networks.} 
    \label{fig:norm}
\end{figure}

Multilingual neural machine translation (MNMT) enables translation between unseen language pairs, i.e., zero-shot translation (ZST)~\cite{johnson-etal-2017-googles,DBLP:journals/csl/FiratCSYB17}. Prior studies have explored techniques such as language tags~\cite{wu-etal-2021-language}, residual connections~\cite{liu-etal-2021-improving-zero}, and novel training objectives~\cite{al-shedivat-parikh-2019-consistency,pham-etal-2019-improving,DBLP:journals/corr/abs-1903-07091,gu-etal-2019-improved,zhu-etal-2020-language,zhang-etal-2020-improving,wang-etal-2021-rethinking-zero,yang-etal-2021-improving-multilingual} for improving ZST. They primarily used the Transformer architecture~\cite{DBLP:conf/nips/VaswaniSPUJGKP17}, which has two variations depending on the position of layer normalization (LayerNorm)~\cite{DBLP:journals/corr/BaKH16}, namely, PreNorm (applied at the input of layers)~\cite{DBLP:conf/iclr/BaevskiA19} and PostNorm (applied after residual connections), as shown in Fig.~\ref{fig:norm}. As previous studies showed that PreNorm can result in more stable training and faster convergence compared to PostNorm for MNMT ~\cite{DBLP:conf/icml/XiongYHZZXZLWL20}, most ZST works~\cite{pham-etal-2019-improving,wu-etal-2021-language,liu-etal-2021-improving-zero} use PreNorm as the default setting following those MNMT studies. However,~\citet{DBLP:conf/nips/Xu0ZZL19} revealed that PreNorm carries the risk of overfitting the training data. We thus hypothesize that in a multilingual scenario, PreNorm may overfit supervised directions and have poor ZST generalizability. We systematically explore PreNorm and PostNorm's effect on ZST to verify this.


Using the OPUS, IWSLT, and Europarl datasets and a total of $54$ ZST directions, we show that PostNorm consistently outperforms PreNorm by up to $12.3$ BLEU points. Following previous work, we also evaluate different language tag~\cite{wu-etal-2021-language} and residual connection~\cite{liu-etal-2021-improving-zero} settings, as they have been shown to impact ZST but we observe that PostNorm continues to be superior thereby lending credibility to our hypothesis.

To better understand the performance differences, we introduce a novel analysis approach called \textbf{layer-wise language recognition} (\textbf{LLR}), which tracks the off-target rates for each encoder and decoder layer by training token-level classifiers to recognize the source or target language. This analysis shows that PreNorm is more sensitive to language tag settings than PostNorm, negatively impacting ZST performance. Additionally, by examining the unraveled view of PreNorm (Fig.~\ref{fig:norm}) inspired by~\citet{DBLP:conf/nips/VeitWB16}, we reveal structural flaws in PreNorm for ZST. Our analysis demonstrates that the order of LayerNorm and self-attention/feed-forward network in PreNorm is the main factor affecting its ZST performance.


Given the prevalent use of PreNorm as the default setting in ZST baselines and frameworks such as Fairseq~\cite{ott-etal-2019-fairseq}\footnote{\url{https://github.com/facebookresearch/fairseq/tree/main/examples/multilingual}} and Tensor2Tensor~\cite{DBLP:conf/amta/VaswaniBBCGGJKK18}, our study emphasizes the importance of careful consideration in the LayerNorm setting for ZST.

\section{Background: LayerNorm}
\label{sec:norm}

LayerNorm~\cite{DBLP:journals/corr/BaKH16} normalizes the input $\mathbf{x}$ by zero-centering and scaling to have a unit standard deviation, followed by an additional trainable transformation, including a gain and bias adjustment. Specifically, it is formulated as:
\begin{equation}
    \text{LayerNorm}(\mathbf{x}) = \frac{\mathbf{x}-\mathbf{E}(\mathbf{x})}{\sqrt{\mathbf{V}(\mathbf{x})}}\cdot\mathbf{g} + \mathbf{b},
\end{equation}
where $\mathbf{g}$ and $\mathbf{b}$ are trainable gain and bias. $\mathbf{E}$ and $\mathbf{V}$ indicate expectation and variance. LayerNorm is commonly used in two positions in the Transformer, as shown in Fig.~\ref{fig:norm}. PostNorm, which is the originally proposed setting of the Transformer~\cite{DBLP:conf/nips/VaswaniSPUJGKP17}, involves applying LayerNorm after each sub-module (i.e., self-attention or feed-forward network) and residual connections. PreNorm~\cite{DBLP:conf/iclr/BaevskiA19}, on the other hand, involves applying LayerNorm directly before each sub-module and is known to stabilize Transformer training. While variants of Transformer LayerNorm like RMSNorm~\cite{DBLP:conf/nips/ZhangS19a} have been proposed, the vanilla PreNorm and PostNorm are still the most widely adopted settings in current multilingual NMT literature. Therefore, we only focus on PreNorm and PostNorm in this work. 

\citet{nguyen-salazar-2019-transformers} have explored the impacts of normalization and initialization choices on supervised low-resource NMT settings, however, we delve deeper and focus on the significance of the positioning of LayerNorm for zero-shot NMT. We expect this to complete the understanding of LayerNorm's role in multilingualism, particularly in the context of zero-shot translation.

\section{Experiments and Results}

We evaluate the performance of PreNorm and PostNorm for ZST on various datasets and language pairs. We then analyze the off-target rates and structural discrepancies between PreNorm and PostNorm to understand performance differences.

\subsection{Experimental Settings}

\begin{table}[t]
    \centering
    \resizebox{\linewidth}{!}{
    \begin{tabular}{llrrl}
        \toprule
        \textbf{Datasets} & \textbf{Languages} & $\mathbf{N}_{zero}$ & $\mathbf{S}_{train}$ & \textbf{Arch.} \\
        \toprule
        \multirow{2}{*}{OPUS} & ar, de, en, & \multirow{2}{*}{30} & \multirow{2}{*}{12.00M} & \multirow{2}{*}{\texttt{base}} \\
        & fr, nl, ru, zh & & & \\
        \hline
        IWSLT & en, it, nl, ro & 6 & 1.38M & \texttt{base} \\
        \hline
        Europarl & de, en, es, fr, nl & 12 & 15.78M & \texttt{big} \\
        \bottomrule
    \end{tabular}
    }
    \caption{\textbf{Statistics of the training data}. $\mathbf{N}_{zero}$ and $\mathbf{S}_{train}$ denote number of the ZST directions and size of the training data, respectively. \texttt{base} and \texttt{big} indicate \texttt{Transformer-base} and \texttt{Transformer-big}.}
    \label{tab:data}
\end{table}

\noindent\textbf{Datasets}
We perform ZST experiments on three datasets: OPUS~\cite{zhang-etal-2020-improving}, IWSLT~\cite{cettolo-etal-2017-overview}, and Europarl~\cite{DBLP:conf/mtsummit/Koehn05}. The statistics of the datasets are summarized in Table~\ref{tab:data}. We include $7$, $4$, and $5$ languages for each dataset. The training data consists of only English-centric sentence pairs, resulting in $30$, $6$, and $12$ ZST directions for each dataset. The total number of parallel sentences for each dataset is $12.00$M, $1.38$M, and $15.78$M, respectively. We apply BPE~\cite{sennrich-etal-2016-neural} with merge operations of $50$k, $40$k, and $50$k to create a joint vocabulary for each dataset.

\noindent\textbf{Training}
We employ \texttt{Transformer-base} model for OPUS and IWSLT, and \texttt{Transformer-big} for Europarl, in accordance with the distinct sizes of training data. We consider the following settings:

\begin{table*}[t]
    \centering
    \resizebox{\linewidth}{!}{
    \begin{tabular}{l|lll|rrr|rrr }
        \toprule
        \multirow{2}{*}{\#} & \textbf{Layer} & Language & \multirow{2}{*}{Res.} & \multicolumn{3}{c|}{Zero-shot} &
        \multicolumn{3}{c}{Supervised} \\
         & \textbf{Norm} & Tag & & OPUS & IWSLT & Europarl & OPUS & IWSLT & Europarl \\
        \toprule
        0 & \multicolumn{3}{c|}{\textit{Pivot}} & 21.8 & 20.0 & 29.5 & - & - & - \\
        \hline
        1 & \textbf{PreNorm} & S-ENC-T-DEC & w/ & 10.1 \small(42.19\%) & 4.9 \small(64.84\%) & 24.9 \small(\textcolor{white}{0}7.73\%) & 33.7 & 31.5 & 34.3 \\
        2 & \textbf{PostNorm} & S-ENC-T-DEC & w/ & \textbf{16.8} \small(\textcolor{white}{0}\textbf{8.59\%}) & \textbf{12.4} \small(\textbf{10.61\%}) & \textbf{29.2} \small(\textcolor{white}{0}\textbf{0.34\%}) & 33.9 & 31.5 & 34.5 \\
        \hline
        3 & \textbf{PreNorm} & T-ENC & w/ & 13.3 \small(22.99\%) & 13.7 \small(\textcolor{white}{0}\textbf{3.98\%}) & 29.5 \small(\textcolor{white}{0}0.23\%) & 33.7 & 31.6 & 34.4 \\
        4 & \textbf{PostNorm} & T-ENC & w/ & \textbf{14.0} \small(\textbf{22.86\%}) & \textbf{15.5} \small(\textcolor{white}{0}4.59\%) & \textbf{30.8} \small(\textcolor{white}{0}\textbf{0.11\%}) & 34.1 & 31.5 & 34.5 \\
        \hline
        5 & \textbf{PreNorm} & S-ENC-T-DEC & w/o & 14.3 \small(20.67\%) & 8.0 \small(50.16\%) & 16.7 \small(41.87\%) & 33.6 & 30.9 & 34.3 \\
        6 & \textbf{PostNorm} & S-ENC-T-DEC & w/o & \textbf{16.0} \small(\textbf{15.27\%}) & \textbf{17.4} \small(\textcolor{white}{0}\textbf{1.83\%}) & \textbf{29.0} \small(\textcolor{white}{0}\textbf{0.41\%}) & 33.8 & 30.7 & 34.4 \\
        \hline
        7 & \textbf{PreNorm} & T-ENC & w/o & 13.4 \small(27.15\%) & 16.2 \small(\textcolor{white}{0}1.54\%) & 29.9 \small(\textcolor{white}{0}2.15\%) & 33.5 & 30.9 & 34.3 \\
        8 & \textbf{PostNorm} & T-ENC & w/o & \textbf{13.9} \small(\textbf{26.68\%}) & \textbf{17.8} \small(\textcolor{white}{0}\textbf{1.50\%}) & \textbf{30.8} \small(\textcolor{white}{0}\textbf{0.13\%}) & 33.9 & 30.6 & 34.4 \\
        \bottomrule
    \end{tabular}
    }
    \caption{\textbf{BLEU scores and off-target rates (shown in brackets)}. We report the average score of three seeds; refer to Appendix~\ref{app:res} for BLEU score of each translation direction and seed. ``Res.'' indicates the residual connection of self-attention in the $4^{th}$ encoder layer. We mark lower off-target rates and significantly higher BLEU scores~\cite{koehn-2004-statistical} between PreNorm and PostNorm in \textbf{bold} for ZST.}
    \label{tab:bleu1}
\end{table*}

\noindent\textbf{(1) PreNorm or PostNorm}:
PreNorm involves LayerNorm directly before each sub-module (i.e., self-attention or feed-forward network), while PostNorm applies LayerNorm after each sub-module and residual connections, as shown in Fig.~\ref{fig:norm}.\footnote{We also experiment with the setting of LayerNorm without trainable parameters~\cite{DBLP:conf/nips/Xu0ZZL19} in Appendix~\ref{app:layernorm-simple}.}

\noindent\textbf{(2) S-ENC-T-DEC or T-ENC}: Source language tag on the encoder-side and target language tag on the decoder-side; or only target language tag on the encoder-side.~\citet{wu-etal-2021-language} showed that this setting impacts ZST for Transformer with PreNorm.

\noindent\textbf{(3) w/ or w/o Res.}: With the residual connection for self-attention in the middle ($4^{th}$) encoder layer or not.~\citet{liu-etal-2021-improving-zero} revealed that ``w/o Res.'' improves ZST for the model trained with PreNorm. We experiment this with different LayerNorm settings as this may reduce the potential of overfitting on supervised directions, then further impacts ZST, which aligns with our hypothesis.

The settings above lead to eight different combinations, shown in Table~\ref{tab:bleu1} (\#1 - \#8). Additional training details are in Appendix~\ref{app:training}.

\subsection{Main Results}

\begin{figure*}[t]
\begin{minipage}[t]{0.33\textwidth}
\begin{center}
\includegraphics[width=\linewidth]{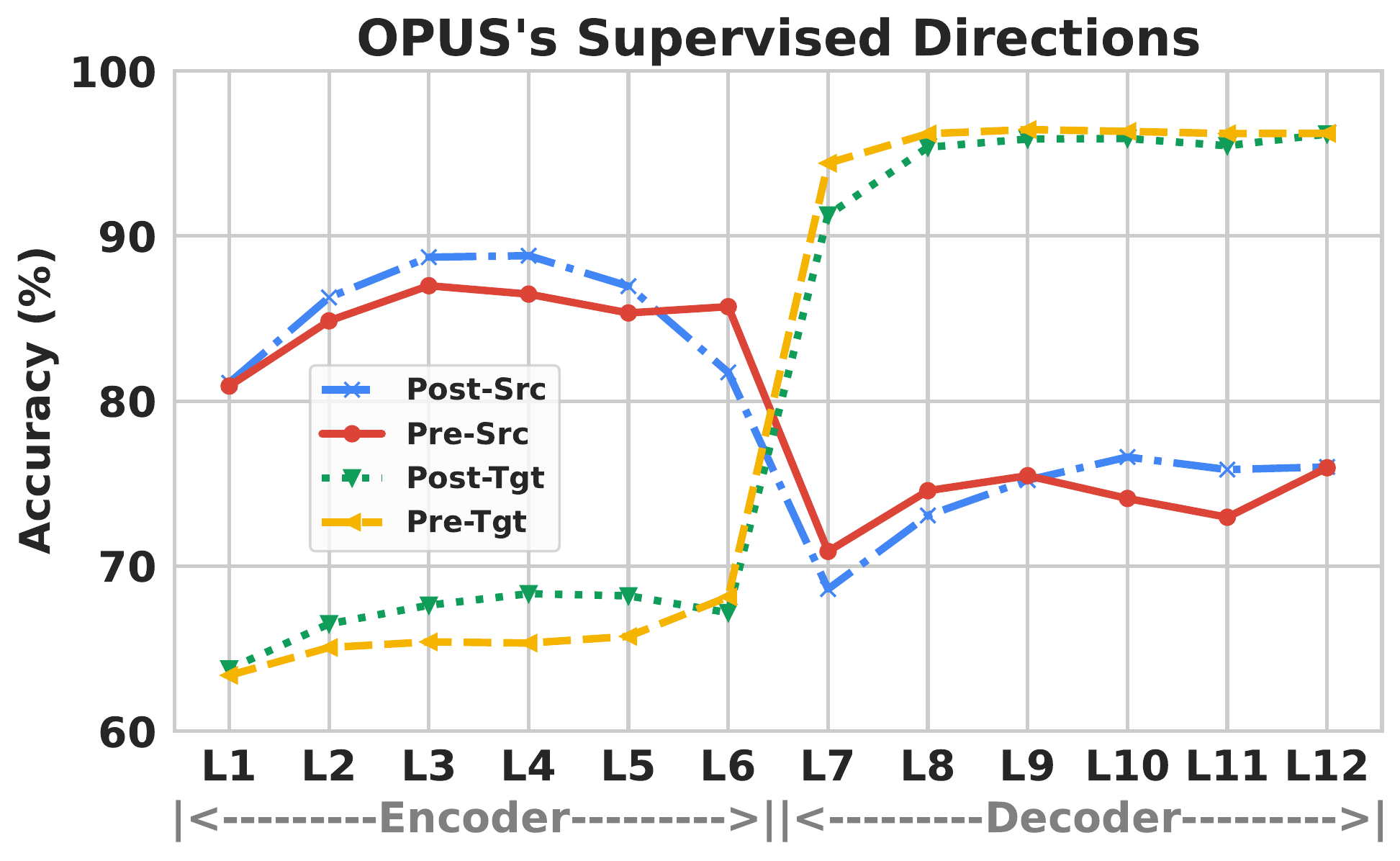}
\end{center}
\end{minipage}
\begin{minipage}[t]{0.33\textwidth}
\begin{center}
\includegraphics[width=\linewidth]{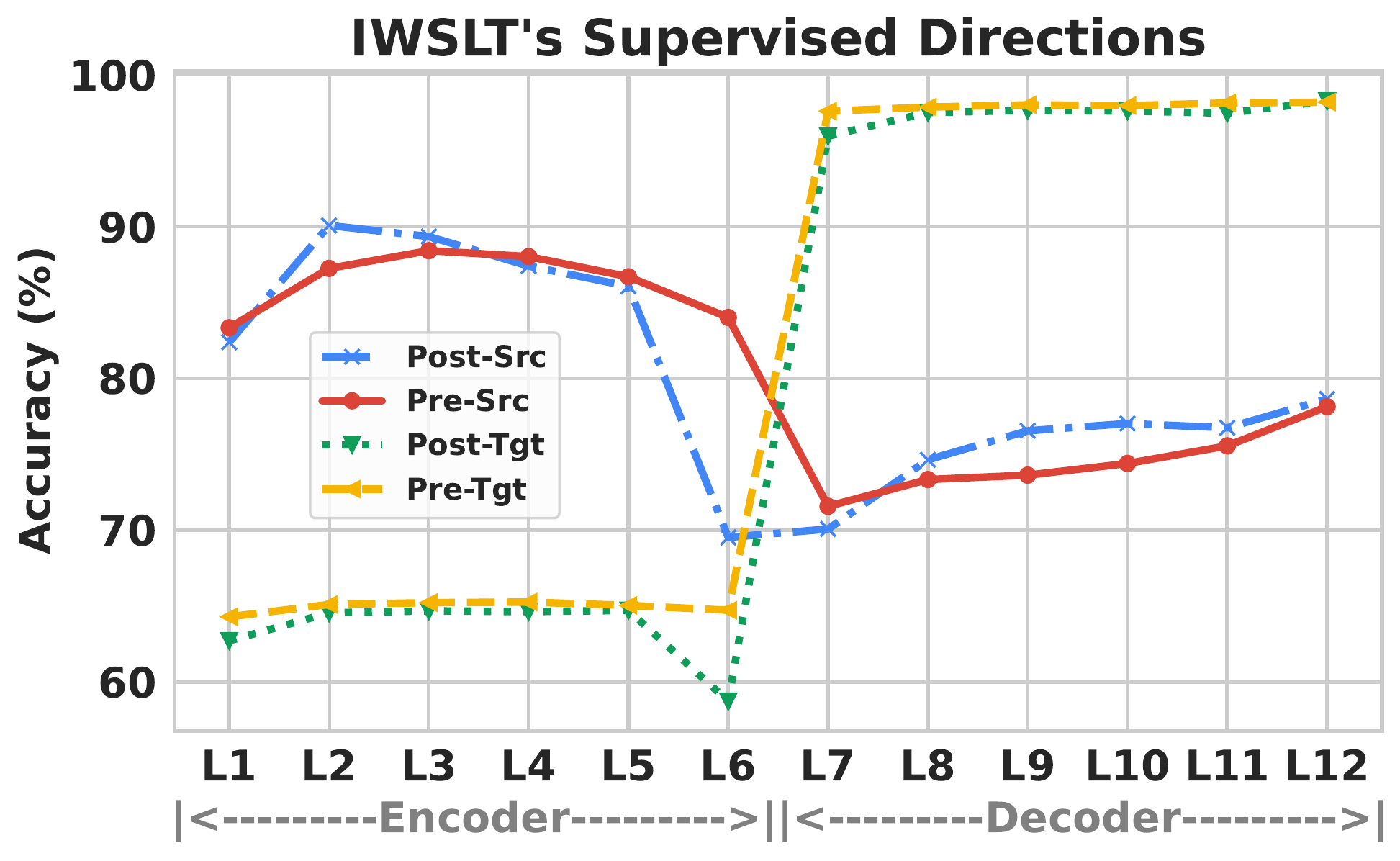}
\end{center}
\end{minipage}
\begin{minipage}[t]{0.33\textwidth}
\begin{center}
\includegraphics[width=\linewidth]{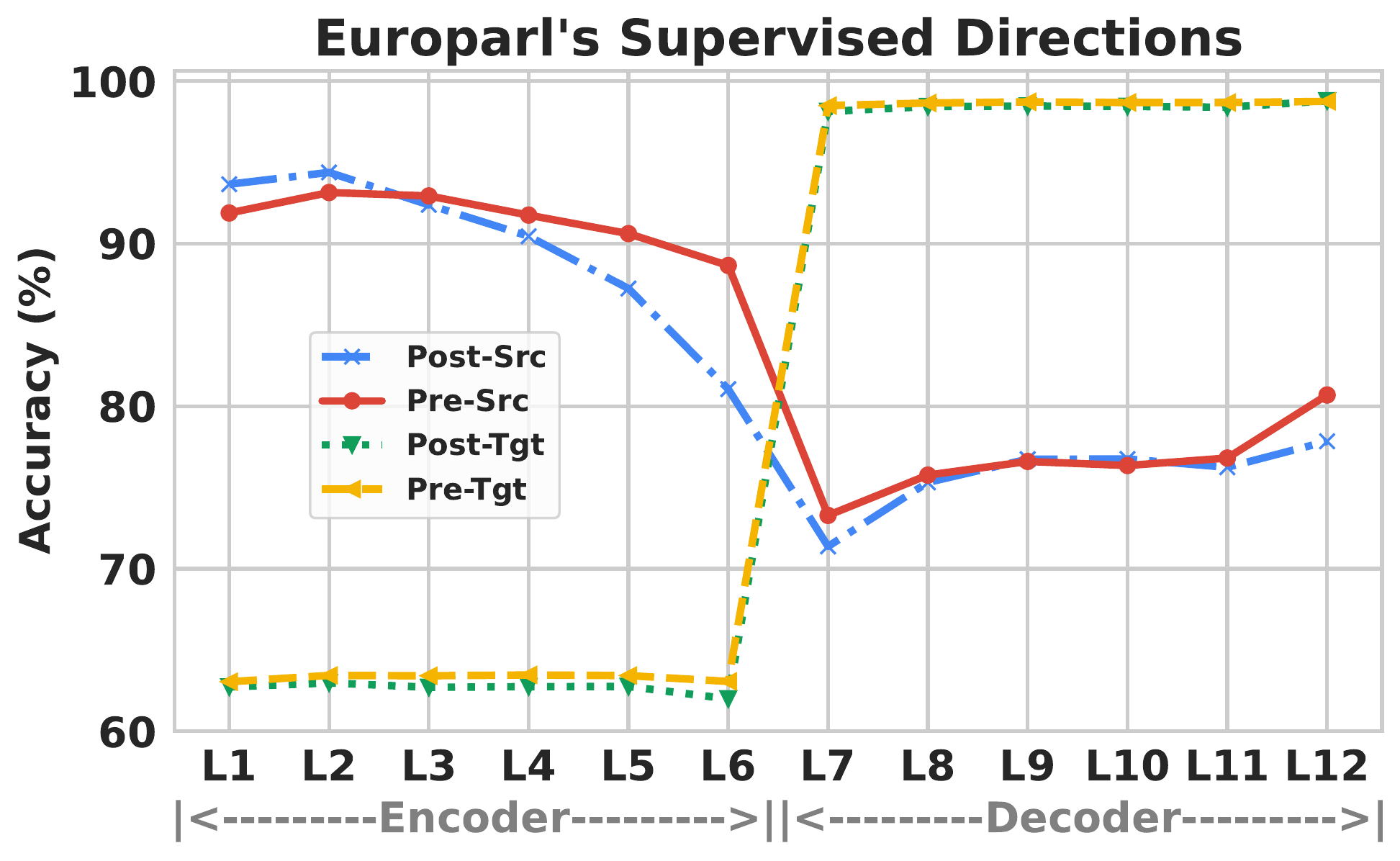}
\end{center}
\end{minipage}
\\
\begin{minipage}[t]{0.33\textwidth}
\begin{center}
\includegraphics[width=\linewidth]{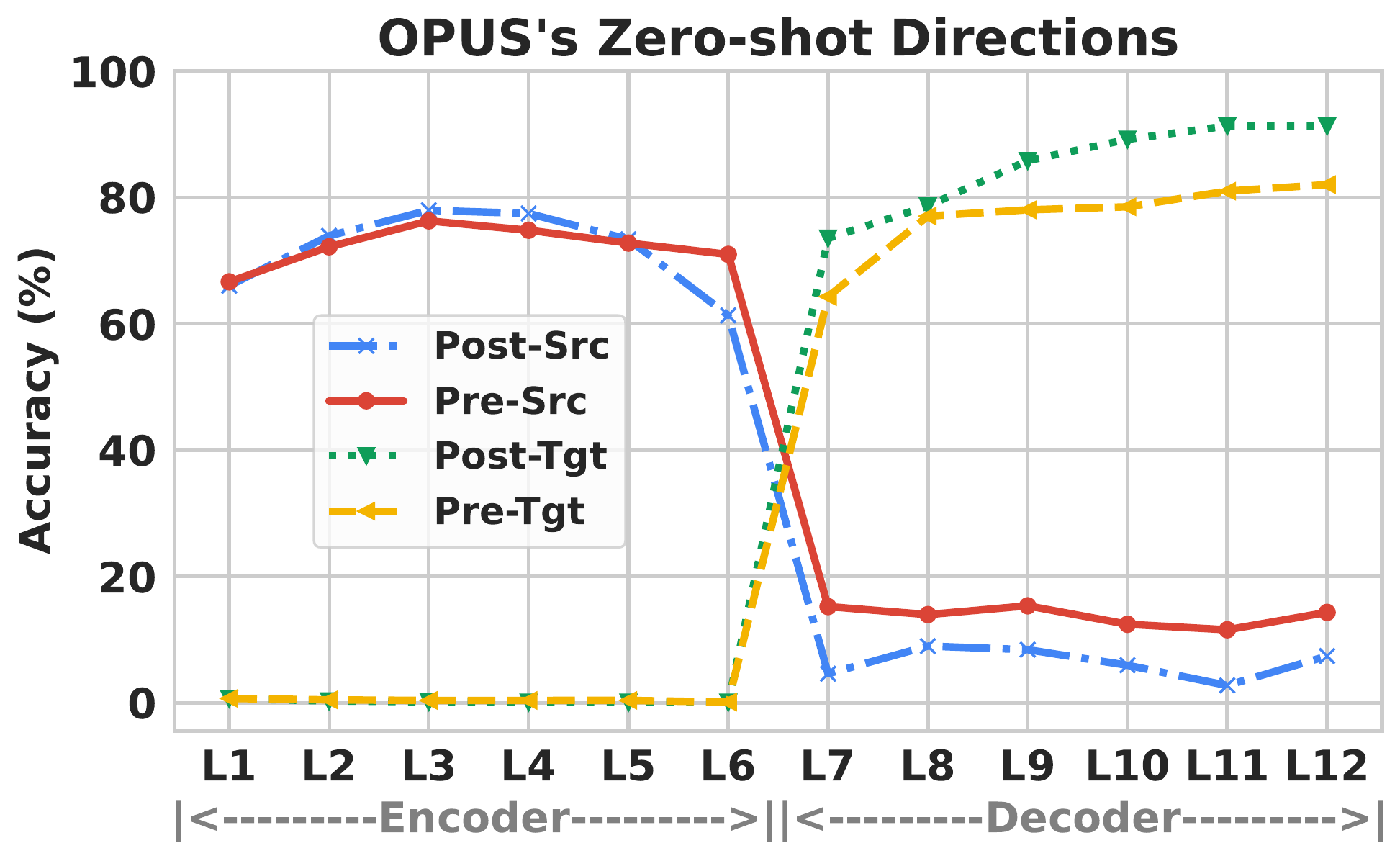}
\end{center}
\end{minipage}
\begin{minipage}[t]{0.33\textwidth}
\begin{center}
\includegraphics[width=\linewidth]{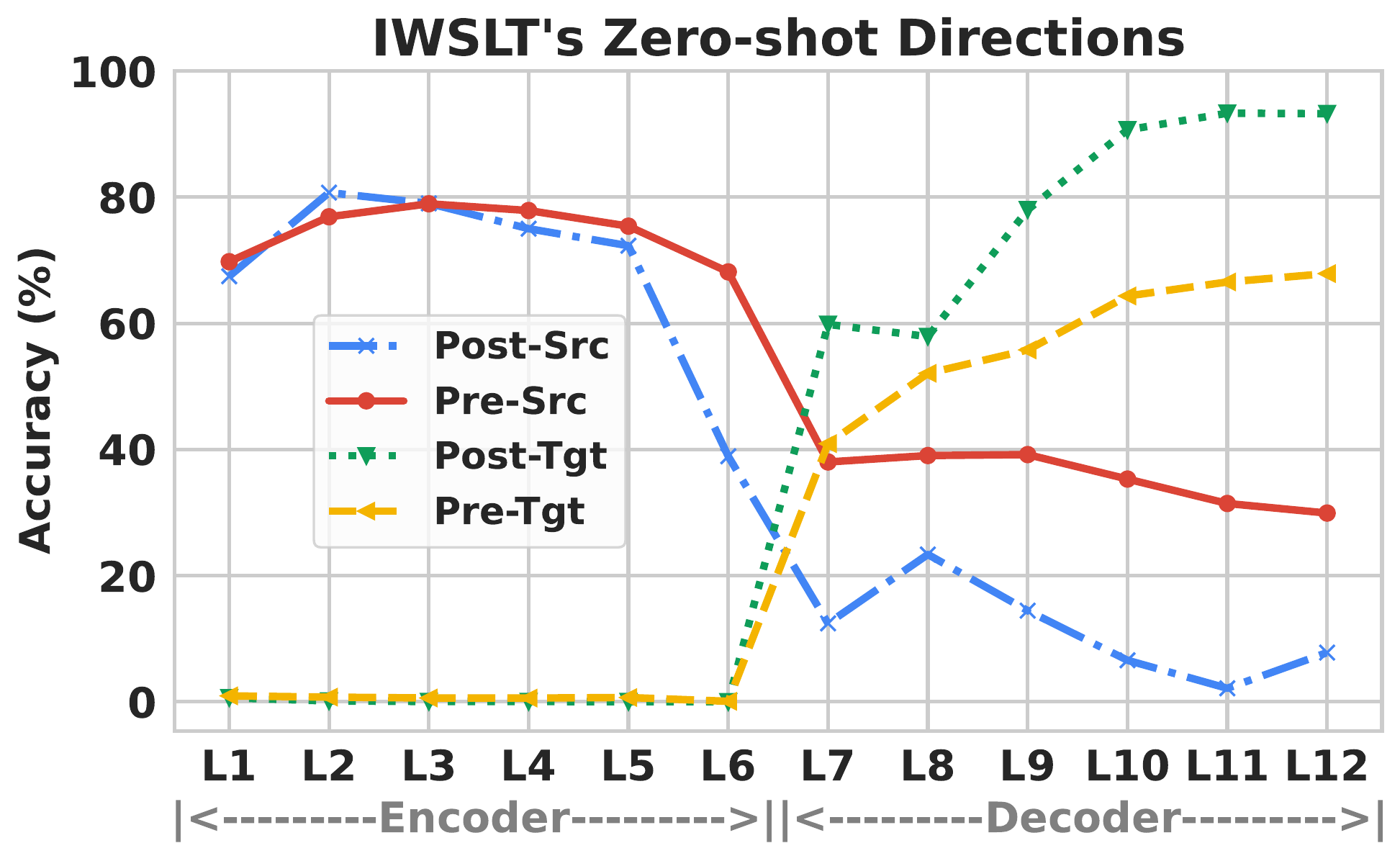}
\end{center}
\end{minipage}
\begin{minipage}[t]{0.33\textwidth}
\begin{center}
\includegraphics[width=\linewidth]{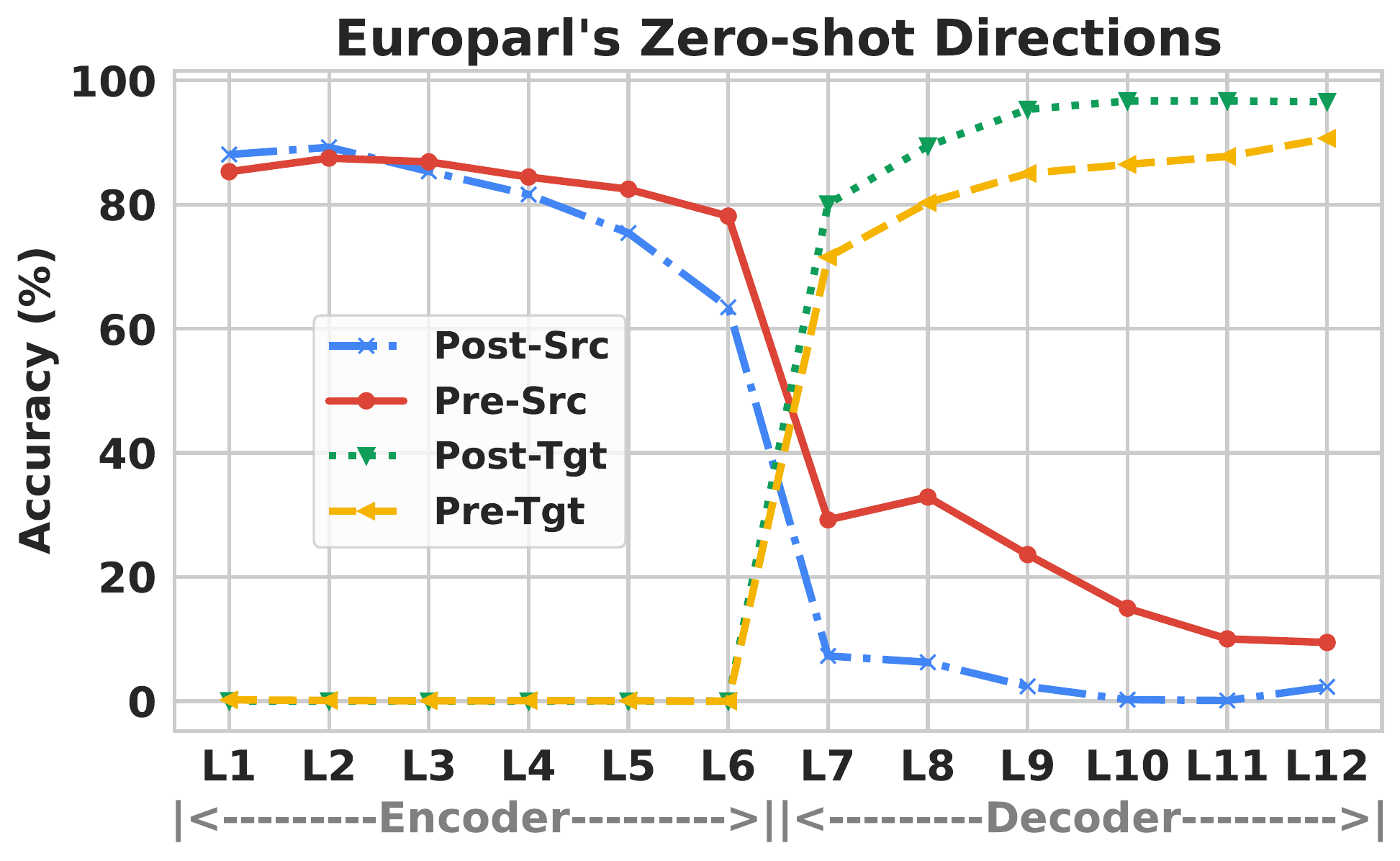}
\end{center}
\end{minipage}
\caption{\textbf{The LLR results of \#1 and \#2 (Table~\ref{tab:bleu1}) for both ZST and supervised directions for each dataset}. We report the average accuracy of three seeds and all the supervised or zero-shot directions. ``Pre-Src'' and ``Pre-Tgt'' indicate the layer-wise source and target language recognition for a PreNorm system (\#1), while ``Post-Src'' and ``Post-Tgt'' denote similary for a PostNorm system (\#2). ``L1'' to ``L6'' are $6$ encoder layers and ``L7'' to ``L12'' are $6$ decoder layers. We present the figures of other systems (\#3 - \#8) in Appendix~\ref{app:llr}.}
\label{fig:off-target}
\end{figure*}

We evaluate ZST systems using SacreBLEU~\cite{post-2018-call} and off-target rates. We report in Table~\ref{tab:bleu1} BLEU scores for both zero-shot and supervised directions. For ZST, we also present pivot-based translation results as a reference. Implementation details of evaluation can be found in Appendix~\ref{app:eval}. Our findings are as follows:

\noindent\textbf{PreNorm vs. PostNorm}:
We find that PostNorm consistently yields better BLEU scores than PreNorm for ZST across various language tag and residual connection settings, while their performance is comparable for supervised directions. 

\noindent\textbf{Impact of Language Tag and Residual Connection:}
We observe that using the ``T-ENC'' language tag and ``w/ Res.'' improves ZST performance for IWSLT, which aligns with the findings of~\citet{wu-etal-2021-language} and~\citet{liu-etal-2021-improving-zero}. Nevertheless, the best performance is achieved using ``w/ Res.'' for PostNorm with ``S-ENC-T-DEC'' and ``T-ENC'' tags for OPUS and Europarl, respectively (\#2 and \#4). Given that~\citet{wu-etal-2021-language} and~\citet{liu-etal-2021-improving-zero} used PreNorm as the default setting (\#2, \#4, \#6 and \#8 are unreported results in their work), our results emphasize the need to consider PostNorm as the default setting for ZST, while the language tag and residual connection settings have less impact.

\noindent\textbf{Off-target Rates}:
Off-target rates help understand the different BLEU score gaps between PreNorm and PostNorm, which ranges from $0.5$ to $12.3$ BLEU points. For PreNorm and PostNorm with the ``T-ENC'' language tag (\#3, \#4, \#7, and \#8), they have similar off-target rates, with a discrepancy ranging from $-0.61\%$ to $2.02\%$, which results in narrow BLEU score gaps, ranging from $0.5$ to $1.8$ points. However, for PreNorm and PostNorm with the ``S-ENC-T-DEC'' language tag (\#1, \#2, \#5, and \#6), the off-target rates show a more considerable discrepancy, ranging from $5.40\%$ to $54.23\%$, resulting in BLEU score gaps from $1.7$ to $12.3$ points. Further analysis of the nature of Transformer hidden states in the next section explores the reason for these different off-target rates in translations.

\subsection{Tracking Off-targets within Transformer}

We probe the language independence of hidden states to track off-targets within Transformer and reveal the differences between PreNorm and PostNorm. In previous work, language independence was primarily analyzed using either SVCCA~\cite{DBLP:conf/nips/RaghuGYS17} or language classification accuracy (LCA)~\cite{liu-etal-2021-improving-zero}. However, we provide evidence in Appendix~\ref{app:svcca} that SVCCA, which measures the cosine similarity between hidden states, are not suitable for ZST systems. Instead, LCA trains a classifier to inspect the hidden states on top of the encoder, but it does not simulate the training of a ZST system, which may introduce bias in the analysis for ZST.\footnote{\citet{liu-etal-2021-improving-zero} regulate the output language via a decoder-side language tag, hence analyzing only the encoder states poses no issues as the target language tag does not impact them. Nevertheless, with other language tag settings such as S-ENC-T-DEC and T-ENC, employed in this study, we require a method to obtain hidden states properly, given their impact on hidden states.} In this work, we propose a novel approach for ZST based on LCA:

\noindent\textbf{LLR} tailors classifiers for each layer to recognize the source or target language. We train a token-level linear classifier for each layer to utilize hidden states in each layer as features to identify the source or target language. We use hidden states obtained by feeding sentence pairs in supervised directions to simulate the training of ZST. We then test each layer's classifer's ability to recognize the source or target language for supervised or zero-shot directions. This approach enables the trained classifier to best represent the language recognition ability of hidden states in a ZST system. 

We train two types of linear classifiers for each encoder and decoder layer. One is for recognizing the source language, and the other is for the target language. Each linear classifier is a linear transformation from the dimension of the hidden states ($512$ or $1,024$) to the number of source or target languages (e.g., $7$ for OPUS). We use the validation set of all supervised directions to obtain the hidden state of each token in each layer and set their source language tag or target language tag as the gold labels. Note that the decoder hidden state of each token in each layer is obtained auto-regressively without teacher-forcing. We train each classifier for $3$ epochs\footnote{The classifier can fully converge within $3$ epochs as the classifier is lightweight that only contains a small number of parameters.} with a learning rate of $1$e-$3$ and a batch size of $64$ sentences. For inference, we utilize the test sets of all supervised or zero-shot directions for computing the LLR results for corresponding directions, respectively.

The LLR results for \#1 and \#2 in Table~\ref{tab:bleu1} are presented in Fig.~\ref{fig:off-target}. First, we find that the encoder and decoder hidden states are highly correlated with the target and source languages, respectively, for supervised directions (L1 to L6 of Pre/Post-Tgt and L7 to L12 of Pre/Post-Src of $3$ upper sub-figures), which may impact the generalizability for ZST. Second, we see that the encoder hidden states of PostNorm are less dependent on the source language than PreNorm (L6 of Pre/Post-Src of $3$ lower sub-figures). Third, we observe that the hidden states in all the decoder layers of PostNorm are more dependent on the target language and less on the source language than PreNorm (L7 to L12 of $3$ lower sub-figures). The latter two points contribute to the observed gaps in off-target rates between PreNorm and PostNorm. Conclusions for \#5 and \#6 with the ``S-ENC-T-DEC'' tag are identical (Appendix~\ref{app:res}). 

For systems using ``T-ENC,'' we find that the LLR are similar between PreNorm and PostNorm (Appendix~\ref{app:res}) and attribute the BLEU score gaps to translation quality (i.e., adequacy and fluency). 

\begin{figure}
    \centering
    \includegraphics[width=\linewidth]{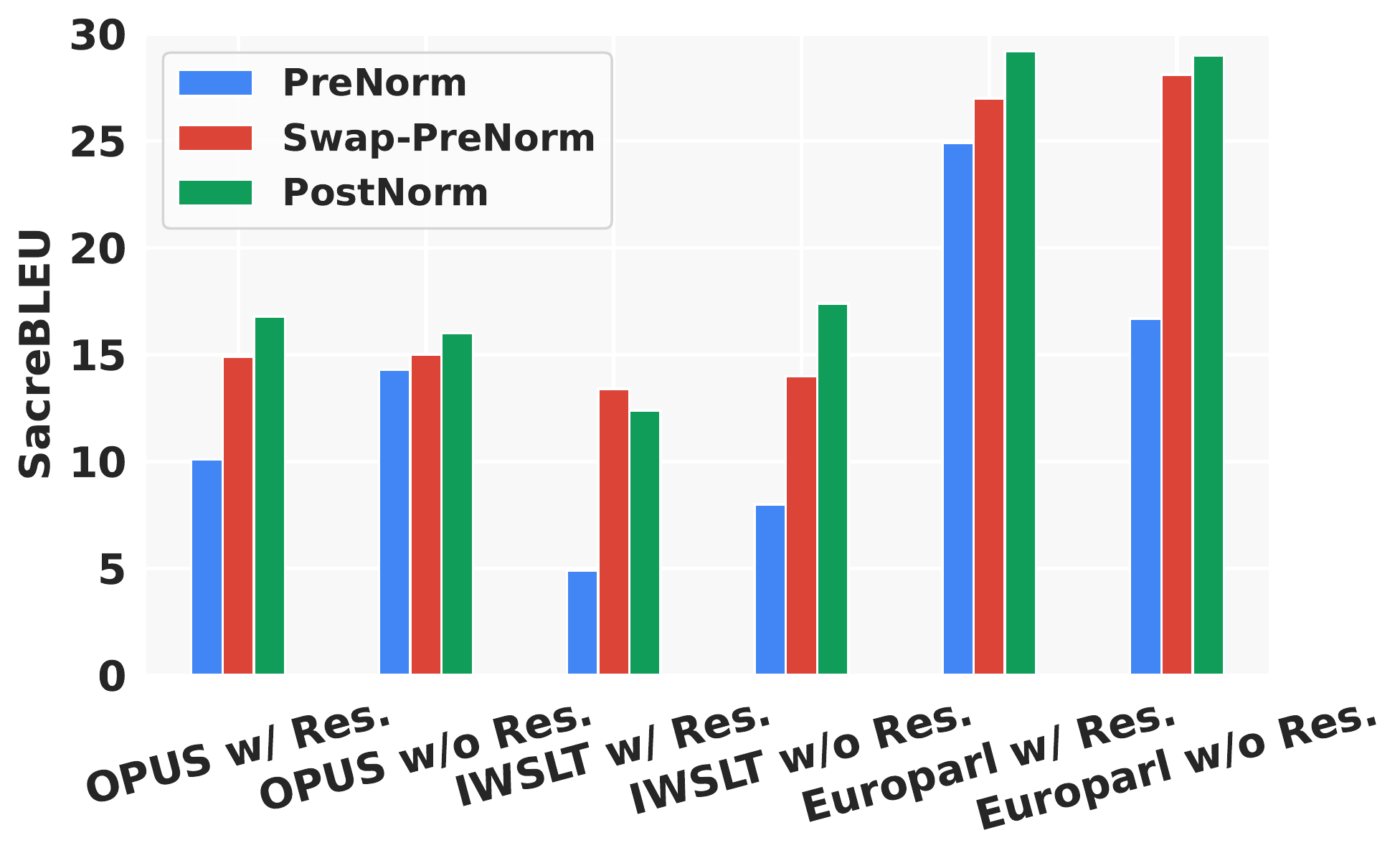}
    \caption{\textbf{BLEU scores of systems with ``S-ENC-T-DEC'' for ZST}. We report the mean of three seeds.}
    \label{fig:swap-prenorm}
\end{figure}

\subsection{Unraveling Structural Flaws of PreNorm}
\label{sec:flaw}

We investigate the structural differences between PreNorm and PostNorm to explain the observed differences in hidden states for models trained with the ``S-ENC-T-DEC'' tag. Inspired by~\citet{DBLP:conf/nips/VeitWB16}, we present an ``unraveled view'' for PreNorm, which decomposes the residual connections by the summation of several sub-networks, as shown in Fig.~\ref{fig:norm} (paths with different colors indicate sub-networks). However, this is not applicable to PostNorm, as LayerNorm is located after residual connections. Based on this analysis, the structural characteristic of PreNorm is:

\noindent\textbf{(1) Shallow Sub-network Nature}: PreNorm includes shallow sub-networks, such as the embedding layer output fed through encoder layers without any operation except for the final LayerNorm (red path in Fig.~\ref{fig:norm}), but PostNorm does not.

\noindent\textbf{(2) LayerNorm Before SA/FFN}: In PreNorm, LayerNorm is placed directly before the self-attention (SA) or feed-forward module (FFN) within the residual connection module. 

To analyze the impact of these structural characteristics on the generalizability of PreNorm in ZST, we swap the order of LayerNorm and SA/FFN within the residual connection module (\textbf{Swap-PreNorm}), while keeping the shallow sub-network nature of PreNorm. Refer to Appendix~\ref{app:swap-norm} for specific illustrations of Swap-PreNorm. The results, presented in Fig~\ref{fig:swap-prenorm}, show that PreNorm can be significantly improved through Swap-PreNorm, with Swap-PreNorm approaching the performance of PostNorm. This demonstrates that ZST is more sensitive to the position of LayerNorm in PreNorm than its shallow sub-network nature.

\section{Conclusion}

In this paper, we comprehensively explored the effects of LayerNorm on ZST performance. Our results demonstrate that PostNorm consistently outperforms PreNorm for ZST, regardless of the language tag and residual connection settings used. Through in-depth analysis of off-target rates and structural flaws in the PreNorm model, we were able to identify the underlying factors that contribute to the performance discrepancy. Our study suggests that care should be taken when selecting the LayerNorm setting for ZST in future research.

\section*{Limitations}
According to us there are 3 limitations of our work which will be addressed in future work.
\begin{itemize}
    \item The impact of LayerNorm, language tags, and residual connection settings on ZST was analyzed in this study. However, other factors, such as the number of layers of the Transformer model, may also have an effect and should be further investigated.
    \item Our conclusions were based on overall scores across all ZST directions. Further examination of how LayerNorm impacts specific language pairs is necessary.
    \item We explored the setting of LayerNorm for ZST systems trained from scratch. Exploration of how the LayerNorm setting of multilingual pre-trained models such as mBART~\cite{liu-etal-2020-multilingual-denoising} impacts the fine-tuning for ZST will be needed.
\end{itemize}

\section*{Ethical Considerations}
In this study, we utilized only publicly accessible datasets for model training. Though our experiments focused on neural machine translation models, it is worth noting that these models may produce biased translations. Although this can be mitigated through a debiasing filtering process, it is beyond the scope of this work. Regarding the composition of this paper, only Grammarly\footnote{\url{https://app.grammarly.com/}} was utilized for grammar correction, and there is no originally machine-generated text in the paper.

\section*{Acknowledgements}
This work was supported by JSPS KAKENHI Grant Number 22KJ1843.

\bibliography{anthology,custom}
\bibliographystyle{acl_natbib}

\appendix

\section{Training Details}
\label{app:training}
For data preprocessing, we utilize jieba\footnote{\url{https://github.com/fxsjy/jieba}} for Chinese segmentation and Moses\footnote{\url{https://github.com/moses-smt/mosesdecoder}}~\cite{koehn-etal-2007-moses} for tokenization of other languages. After applying BPE, we obtain vocabularies with sizes of $66,158$, $40,100$, and $50,363$ for OPUS, IWSLT, and Europarl, respectively. For multilingual training, we do not apply oversampling as the data size for each language pair is comparable. The maximum sentence length is set to $256$. We train Transformer models using Fairseq\footnote{\url{https://github.com/facebookresearch/fairseq}} and set the dropout rate to $0.1$, $0.4$, and $0.3$ for each dataset. Adam~\cite{DBLP:journals/corr/KingmaB14} is used as the optimizer with a learning rate of $5$e-$4$, $1$e-$3$, and $5$e-$4$ for each dataset, and $4,000$ warm-up steps are employed. We train the Transformer-base model using $4$ 32G V100 GPUs and the Transformer-big model using $8$ 32G V100 GPUs with the batch size of $4,096$ tokens. Additionally, we employ mixed precision training~\cite{DBLP:conf/iclr/MicikeviciusNAD18} to accelerate the training process. We train each dataset for 200, 100, and 400 epochs, respectively.

\section{Evaluation Details}
\label{app:eval}
For OPUS, we use the test sets following~\cite{zhang-etal-2020-improving}, while for IWSLT and Europarl, we choose the test sets following~\cite{wu-etal-2021-language}. We select the checkpoint with the lowest validation loss for evaluation. The inference is performed on the trained models using a beam size of $5$. For calculating SacreBLEU,\footnote{\url{https://github.com/mjpost/sacrebleu}} we utilize the ``zh'' tokenization mode for Chinese, and the ``13a'' tokenization mode for other languages. We use the model of setting \#4\footnote{We use this setting as it achieves the best performance for supervised directions, as shown in Table~\ref{tab:bleu1}.} (Table~\ref{tab:bleu1}) for pivot-based translation. To calculate the off-target rates, we utilize the language identification tool provided by FastText~\cite{DBLP:journals/corr/JoulinGBDJM16}.\footnote{\url{https://fasttext.cc/docs/en/language-identification.html}} Our experiment has revealed that this tool is slightly more accurate than another tool called ``langdetect,''\footnote{\url{https://github.com/Mimino666/langdetect}} as it can achieve an accuracy of $98\%$ when decoding reference English sentences in the test set, whereas ``langdetect'' only achieves accuracy of around $92\%$.

\begin{table}[t]
    \centering
    \resizebox{\linewidth}{!}{
    \begin{tabular}{lrr}
        \toprule
         & Zero-shot & Supervised \\
        \toprule
        PreNorm & 9.8 & 33.8 \\
        PostNorm & 17.5 & 33.8 \\
        PreNorm w/o Enc-Last & 11.2 & 33.7 \\
        \bottomrule
    \end{tabular}
    }
    \caption{\textbf{BLEU scores of PreNorm, PostNorm, and ``PreNorm w/o Enc-Last'' on OPUS}. They are trained with the ``S-ENC-T-DEC'' tag, ``Res.,'' and the random seed of 10. We report the mean of all the translation directions.}
    \label{tab:prenorm-wo-enclast}
\end{table}

\begin{figure}[t]
    \centering
    \includegraphics[width=\linewidth]{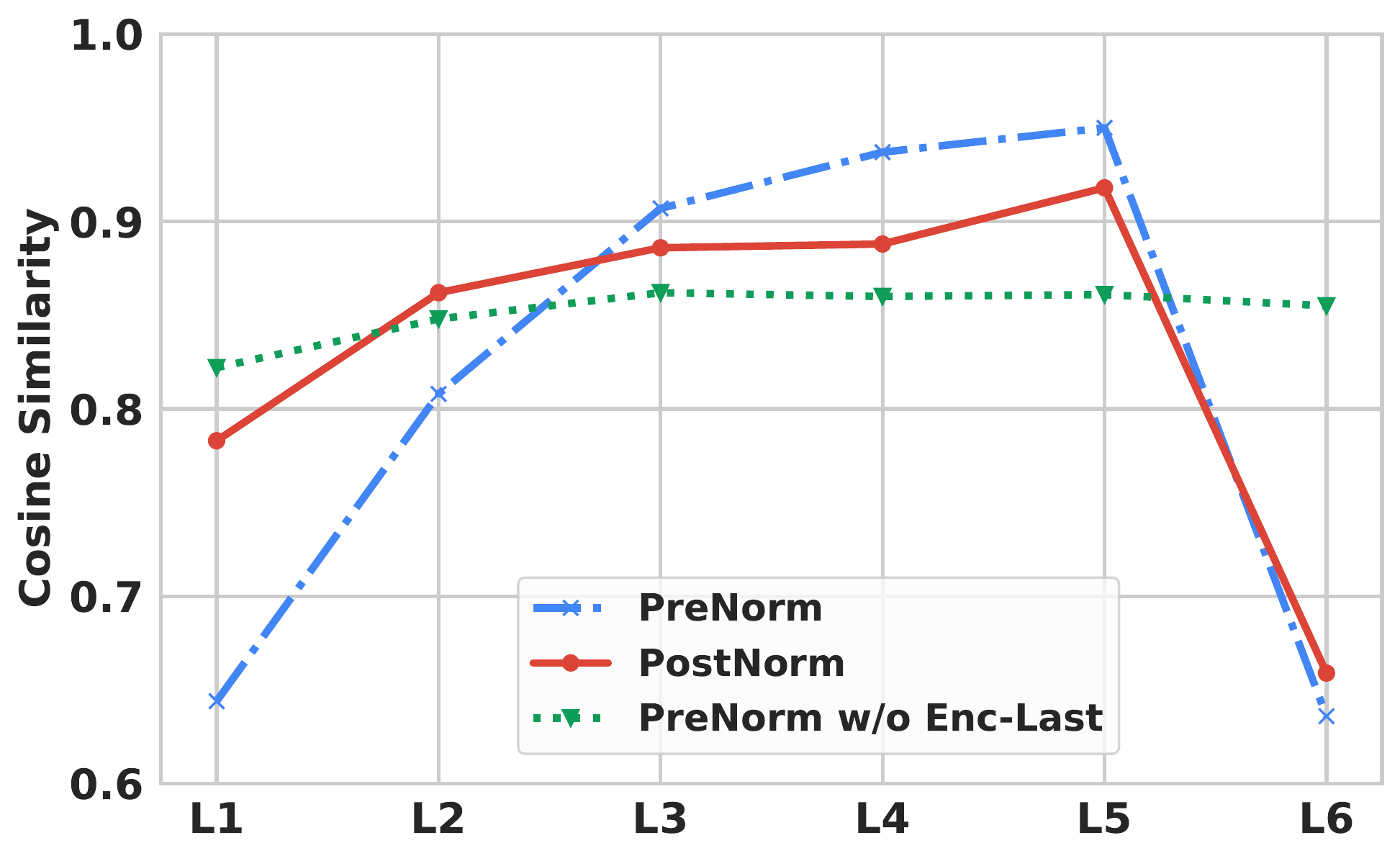}
    \caption{\textbf{Encoder layer-wise SVCCA scores of PreNorm, PostNorm, and ``PreNorm w/o Enc-Last'' between ``en-xx'' and ``xx-en'' translation directions}. We report the mean of all the direction pairs.}
    \label{fig:svcca}
\end{figure}

\begin{figure}[t]
    \centering
    \includegraphics[width=\linewidth]{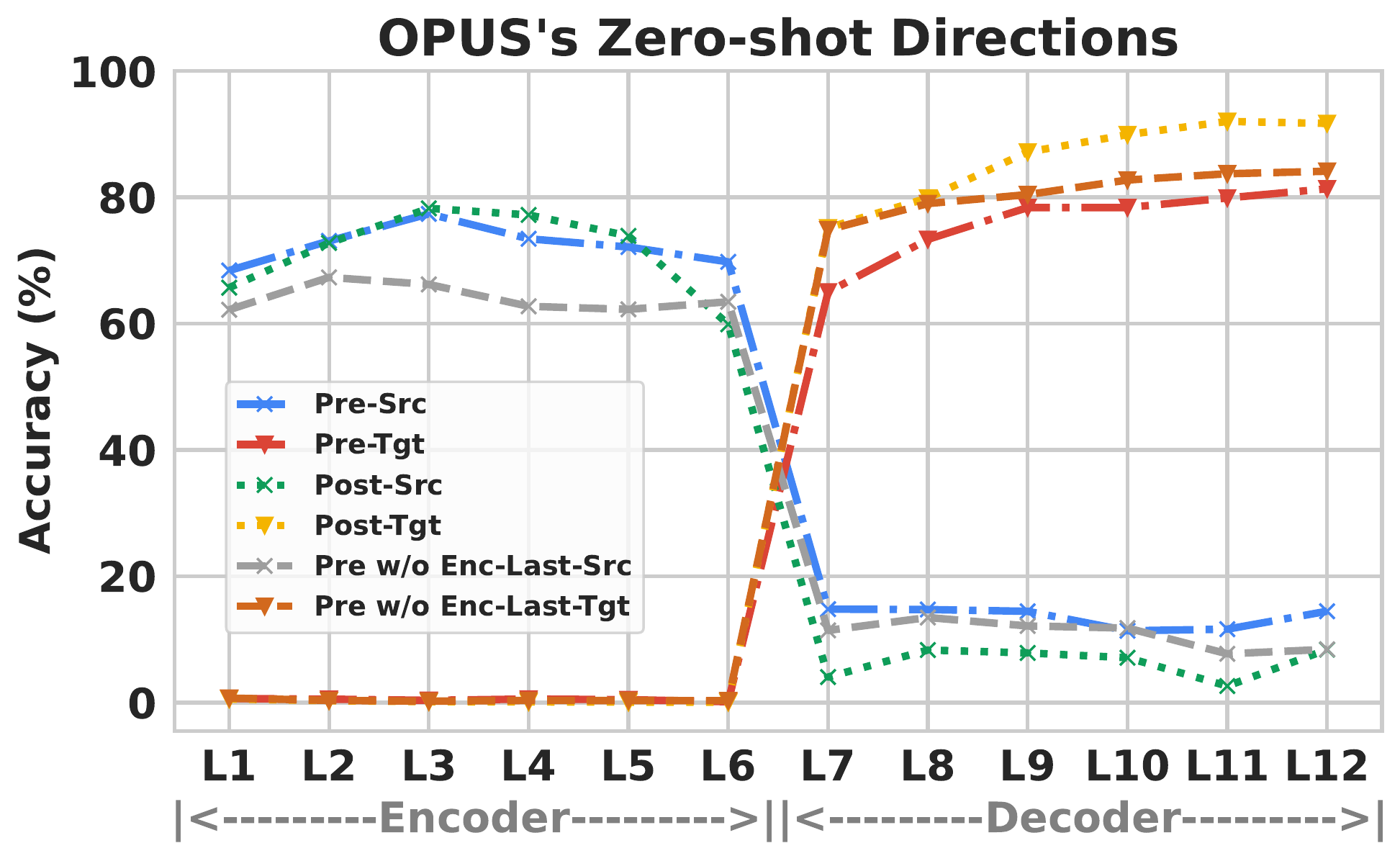}
    \caption{\textbf{The LLR results of PreNorm, PostNorm, and ``PreNorm w/o Enc-Last.''} We report the mean of all the ZST directions. ``-Src'' and ``-Tgt'' indicate the LLR results for the source and target languages, respectively. ``L1'' to ``L6'' are $6$ encoder layers and ``L7'' to ``L12'' are $6$ decoder layers.}
    \label{fig:llr}
\end{figure}

\section{Discussion about SVCCA score}
\label{app:svcca}
In previous work~\cite{wu-etal-2021-language,liu-etal-2021-improving-zero}, the SVCCA score~\cite{DBLP:conf/nips/RaghuGYS17}, a cosine similarity measure between the hidden states of neural models, was used to compare two ZST models. However, we demonstrate that this method is unsuitable for comparing different ZST systems through an experiment. We removed the final LayerNorm from the PreNorm encoder, denoting it as ``PreNorm w/o Enc-Last.'' We then evaluated the BLEU scores of PreNorm, PostNorm, and ``PreNorm w/o Enc-Last'' on the OPUS dataset, as reported in Table~\ref{tab:prenorm-wo-enclast}. We subsequently calculated the encoder layer-wise SVCCA score for each LayerNorm setting using the mean-pooled hidden states of each encoder layer. The average SVCCA score between all the ``en-xx'' and ``xx-en'' directions is reported in Fig.~\ref{fig:svcca}. When comparing Fig.~\ref{fig:svcca} with Table~\ref{tab:prenorm-wo-enclast}, we observe that PostNorm has a higher SVCCA score on top of the encoder (L6) than PreNorm, which suggests that the encoder of PostNorm is more language-agnostic and thus has a higher ZST BLEU score in Table~\ref{tab:prenorm-wo-enclast}, aligning with the results found in~\citet{wu-etal-2021-language} and~\citet{liu-etal-2021-improving-zero}. However, ``PreNorm w/o Enc-Last'' shows an extremely high SVCCA score on top of the encoder, whereas its ZST BLEU performance is significantly lower than PostNorm by $6.3$ BLEU points. This reveals the significant inconsistency between the SVCCA score and the performance of ZST models. Therefore, it is crucial to carefully consider how to leverage SVCCA for ZST analysis in the future.

On the other hand, our proposed LLR score is consistent with the ZST BLEU score, as shown in Fig.~\ref{fig:llr}. Specifically, we observe the lowest LLR score on top of the encoder of PostNorm for the source language and the highest LLR scores in all the decoder layers, which aligns with its best ZST performance among the three systems.

\begin{figure}[t]
    \centering
    \includegraphics[width=\linewidth]{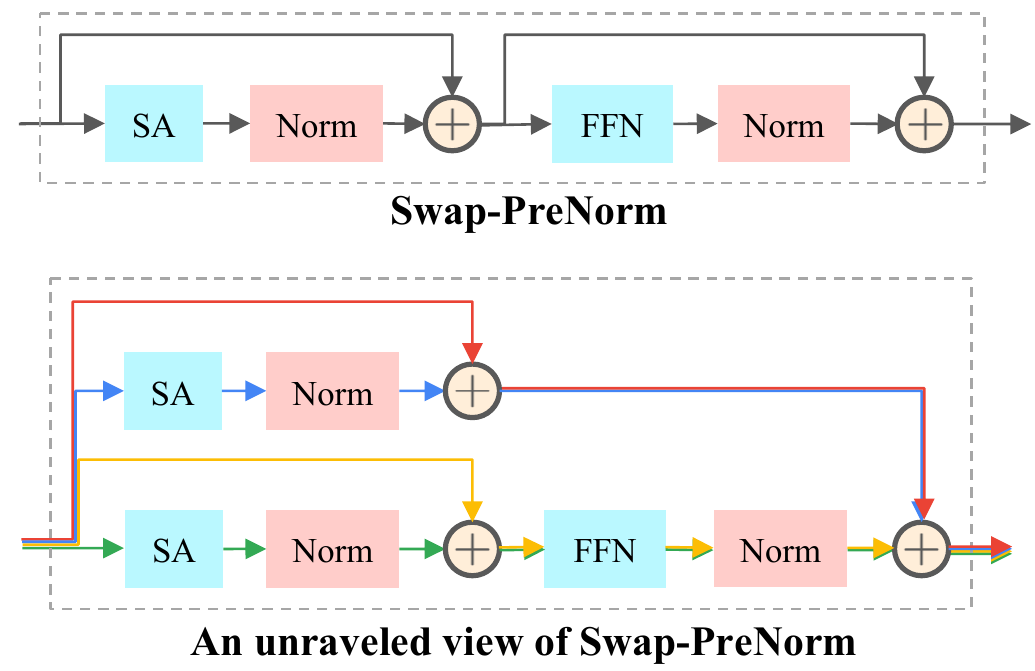}
    \caption{\textbf{Swap-PreNorm, and an unraveled view of Swap-PreNorm in a Transformer encoder layer}. ``Norm,'' ``SA,'' and ``FFN'' denote LayerNorm, self-attention, and feed-forward network. $\oplus$ is residual connection. Paths with different colors in the unraveled view of PreNorm indicate respective sub-networks.}
    \label{fig:swap-prenorm-arch}
\end{figure}

\begin{table*}[t]
    \centering
    \resizebox{\linewidth}{!}{
    \begin{tabular}{l|lll|rrr|rrr }
        \toprule
        \multirow{2}{*}{\#} & \textbf{LayerNorm-} & Language & \multirow{2}{*}{Res.} & \multicolumn{3}{c|}{Zero-shot} &
        \multicolumn{3}{c}{Supervised} \\
         & \textbf{simple} & Tag & & OPUS & IWSLT & Europarl & OPUS & IWSLT & Europarl \\
        \toprule
        1 & \textbf{PreNorm-simple} & S-ENC-T-DEC & w/ & 10.1 \small(+0.0) & 5.9 \small(\textcolor{blue}{+1.0}) & 25.0 \small(+0.1) & 33.9 \small(+0.2) & 31.9 \small(+0.4) & 34.4 \small(+0.1) \\
        2 & \textbf{PostNorm-simple} & S-ENC-T-DEC & w/ & \textbf{15.8} \small(\textcolor{red}{-1.0}) & \textbf{11.5} \small(\textcolor{red}{-0.9}) & \textbf{28.7} \small(\textcolor{red}{-0.5}) & 34.1 \small(+0.2) & 32.1 \small(\textcolor{blue}{+0.6}) & 34.5 \small(+0.0) \\
        \hline
        3 & \textbf{PreNorm-simple} & T-ENC & w/ & 13.7 \small(+0.4) & 14.5 \small(\textcolor{blue}{+0.8}) & 29.4 \small(-0.1) & 33.9 \small(+0.2) & 31.9 \small(+0.3) & 34.4 \small(+0.0) \\
        4 & \textbf{PostNorm-simple} & T-ENC & w/ & \textbf{14.9} \small(\textcolor{blue}{+0.9}) & \textbf{15.4} \small(-0.1) & \textbf{30.8} \small(+0.0) & 34.0 \small(-0.1) & 31.9 \small(+0.4) & 34.6 \small(+0.1) \\
        \hline
        5 & \textbf{PreNorm-simple} & S-ENC-T-DEC & w/o & 15.4 \small(\textcolor{blue}{+1.1}) & 7.8 \small(-0.2) & 19.4 \small(\textcolor{blue}{+2.7}) & 33.7 \small(+0.1) & 31.3 \small(+0.4) & 34.1 \small(-0.2) \\
        6 & \textbf{PostNorm-simple} & S-ENC-T-DEC & w/o & \textbf{16.4} \small(+0.4) & \textbf{16.0} \small(\textcolor{red}{-1.4}) & \textbf{29.2} \small(+0.2) & 33.9 \small(+0.1) & 31.3 \small(\textcolor{blue}{+0.6}) & 34.4 \small(+0.0) \\
        \hline
        7 & \textbf{PreNorm-simple} & T-ENC & w/o & 13.1 \small(-0.3) & 16.8 \small(\textcolor{blue}{+0.6}) & 28.7 \small(\textcolor{red}{-1.2}) & 33.7 \small(+0.2) & 31.4 \small(\textcolor{blue}{+0.5}) & 34.3 \small(+0.0) \\
        8 & \textbf{PostNorm-simple} & T-ENC & w/o & \textbf{14.0} \small(+0.1) & \textbf{17.9} \small(+0.1) & \textbf{31.0} \small(+0.2) & 33.7 \small(-0.2) & 31.1 \small(\textcolor{blue}{+0.5}) & 34.4 \small(+0.0) \\
        \bottomrule
    \end{tabular}
    }
    \caption{\textbf{BLEU scores of LayerNorm-simple}. We report the average score of three seeds. ``Res.'' indicates the residual connection of self-attention in the $4^{th}$ encoder layer. We mark better scores between PreNorm-simple and PostNorm-simple in \textbf{bold}. For each setting, significantly \textcolor{blue}{better} or \textcolor{red}{worse} BLEU scores~\cite{koehn-2004-statistical} compared with the results in Table~\ref{tab:bleu1} are marked in \textcolor{blue}{blue} or \textcolor{red}{red}.}
    \label{tab:bleu2}
\end{table*}

\section{Swap-PreNorm}
\label{app:swap-norm}
Fig.~\ref{fig:swap-prenorm-arch} illustrates the implementation of Swap-PreNorm, which incorporates LayerNorm following the SA/FFN layers within the residual connection block. Compared with PostNorm, Swap-PreNorm alters the order of LayerNorm and residual connections. As depicted in the unraveled view of Swap-PreNorm in Fig.~\ref{fig:swap-prenorm-arch}, it preserves the shallow sub-network characteristics of PreNorm, which is the main difference compared with PostNorm.

\section{LayerNorm without Trainable Parameters}
\label{app:layernorm-simple}
\citet{DBLP:conf/nips/Xu0ZZL19} demonstrated that the overfitting issue of PreNorm can be alleviated by removing the trainable parameters of LayerNorm (LayerNorm-simple). We apply this technique to our ZST experimental settings to investigate the overfitting state of PreNorm and PostNorm. PreNorm and PostNorm after applying this technique are denoted as PreNorm-simple and PostNorm-simple. As reported in Table~\ref{tab:bleu2}, the results indicate that PreNorm-simple and PostNorm-simple outperform their respective original versions in supervised directions, which aligns with the findings of~\citet{DBLP:conf/nips/Xu0ZZL19}. Additionally, we observe comparable or better BLEU scores for PreNorm-simple than PreNorm (except for \#7 on Europarl), indicating that the original PreNorm had low generalizability for ZST. For PostNorm-simple, we observe significant improvement only for \#4 on OPUS, which suggests the superior generalizability of the original PostNorm for ZST. Despite this improvement, PreNorm-simple still underperforms PostNorm, highlighting the severe overfitting problem of the original PreNorm.

\section{Details of the LLR Results}
\label{app:llr}
\begin{figure*}[t]
\begin{minipage}[t]{0.33\textwidth}
\begin{center}
\includegraphics[width=\linewidth]{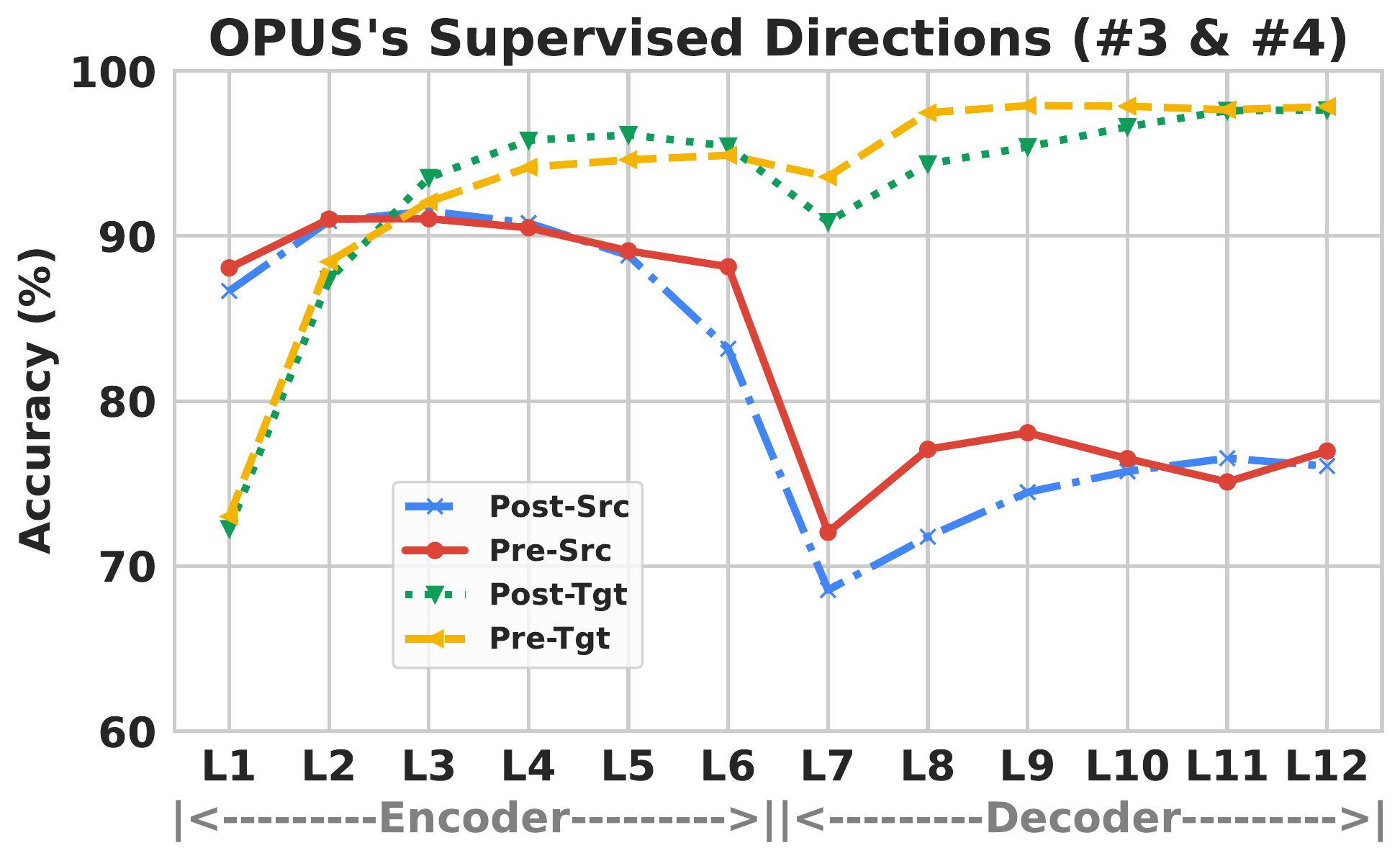}
\end{center}
\end{minipage}
\begin{minipage}[t]{0.33\textwidth}
\begin{center}
\includegraphics[width=\linewidth]{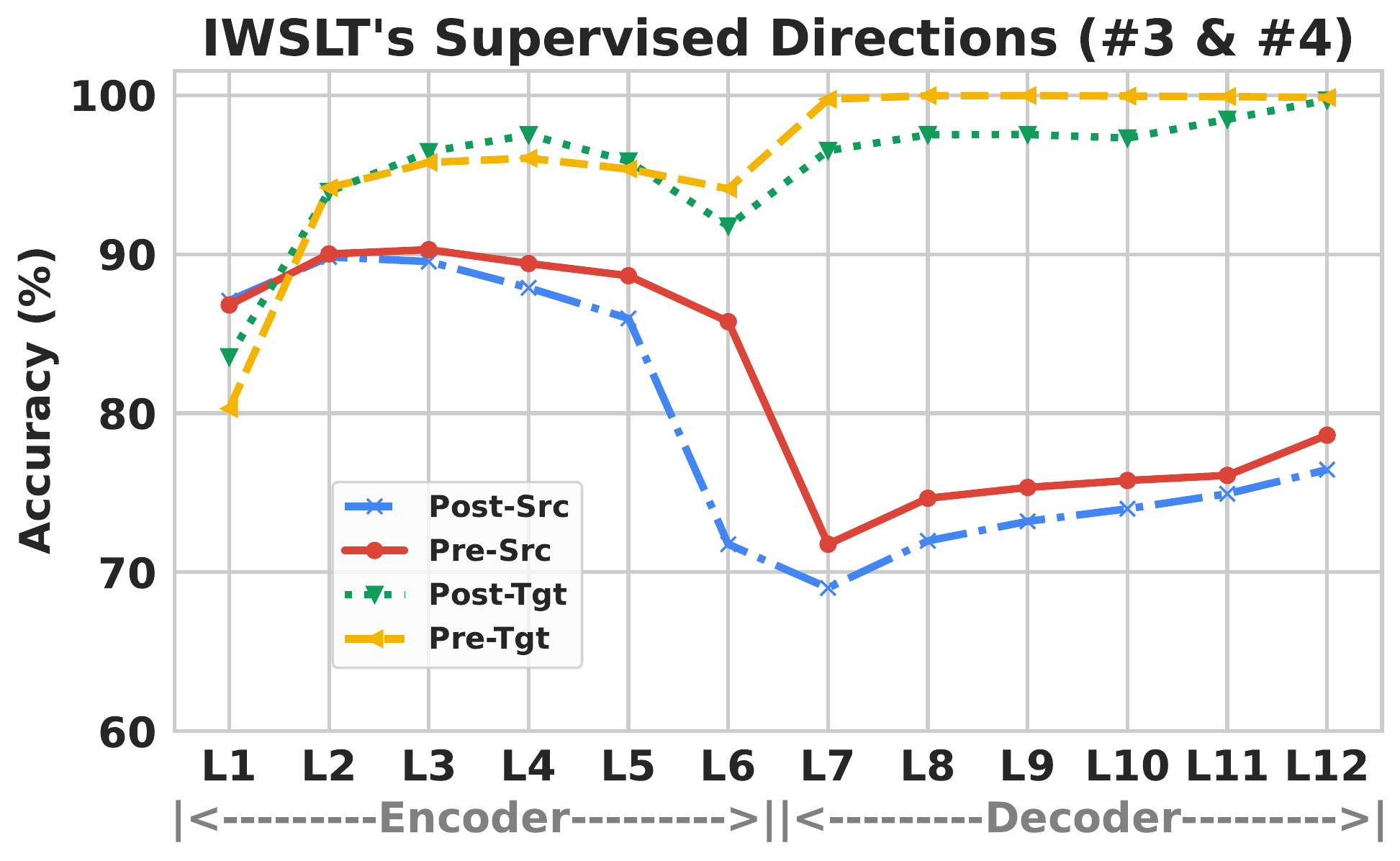}
\end{center}
\end{minipage}
\begin{minipage}[t]{0.33\textwidth}
\begin{center}
\includegraphics[width=\linewidth]{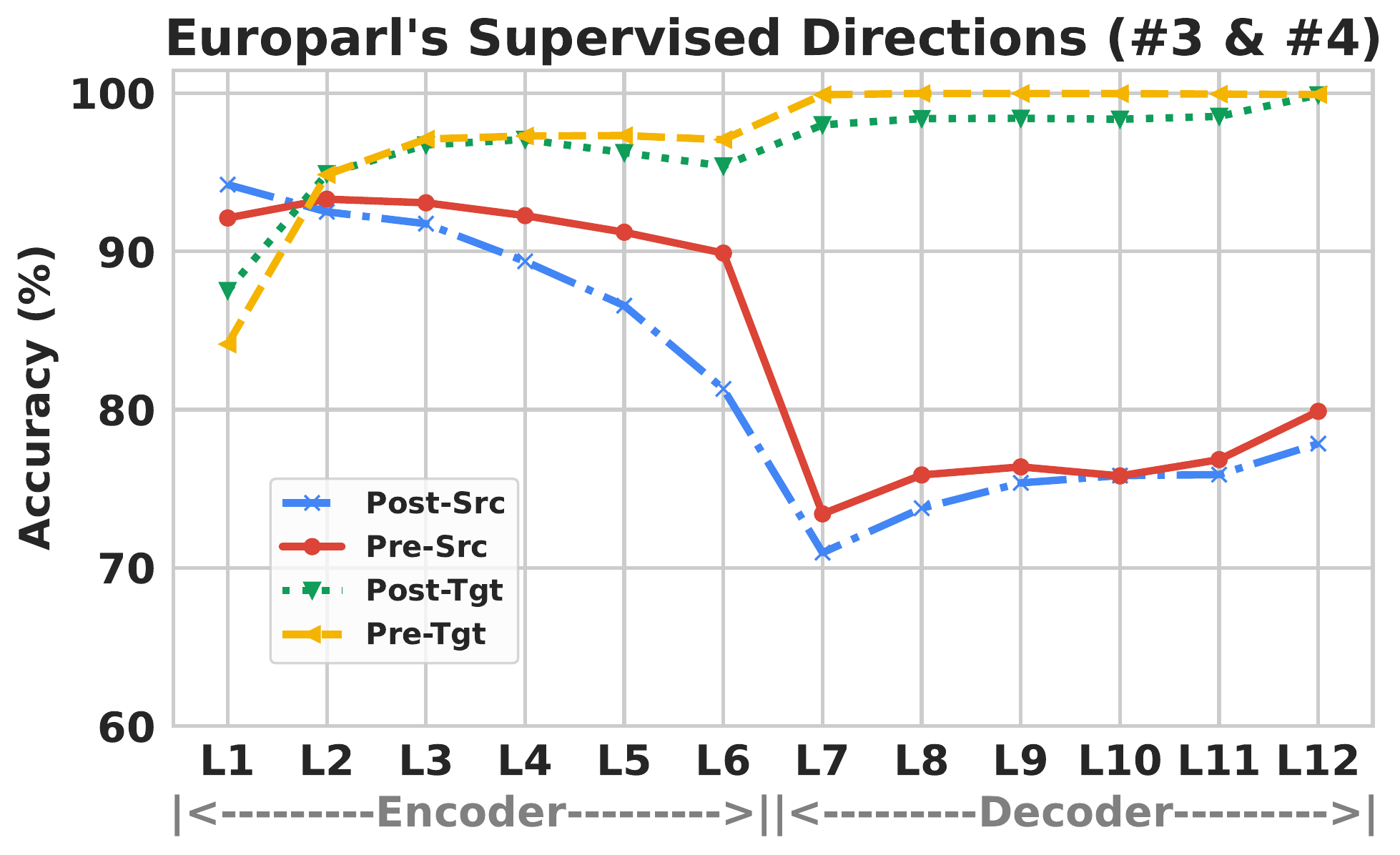}
\end{center}
\end{minipage}
\\
\begin{minipage}[t]{0.33\textwidth}
\begin{center}
\includegraphics[width=\linewidth]{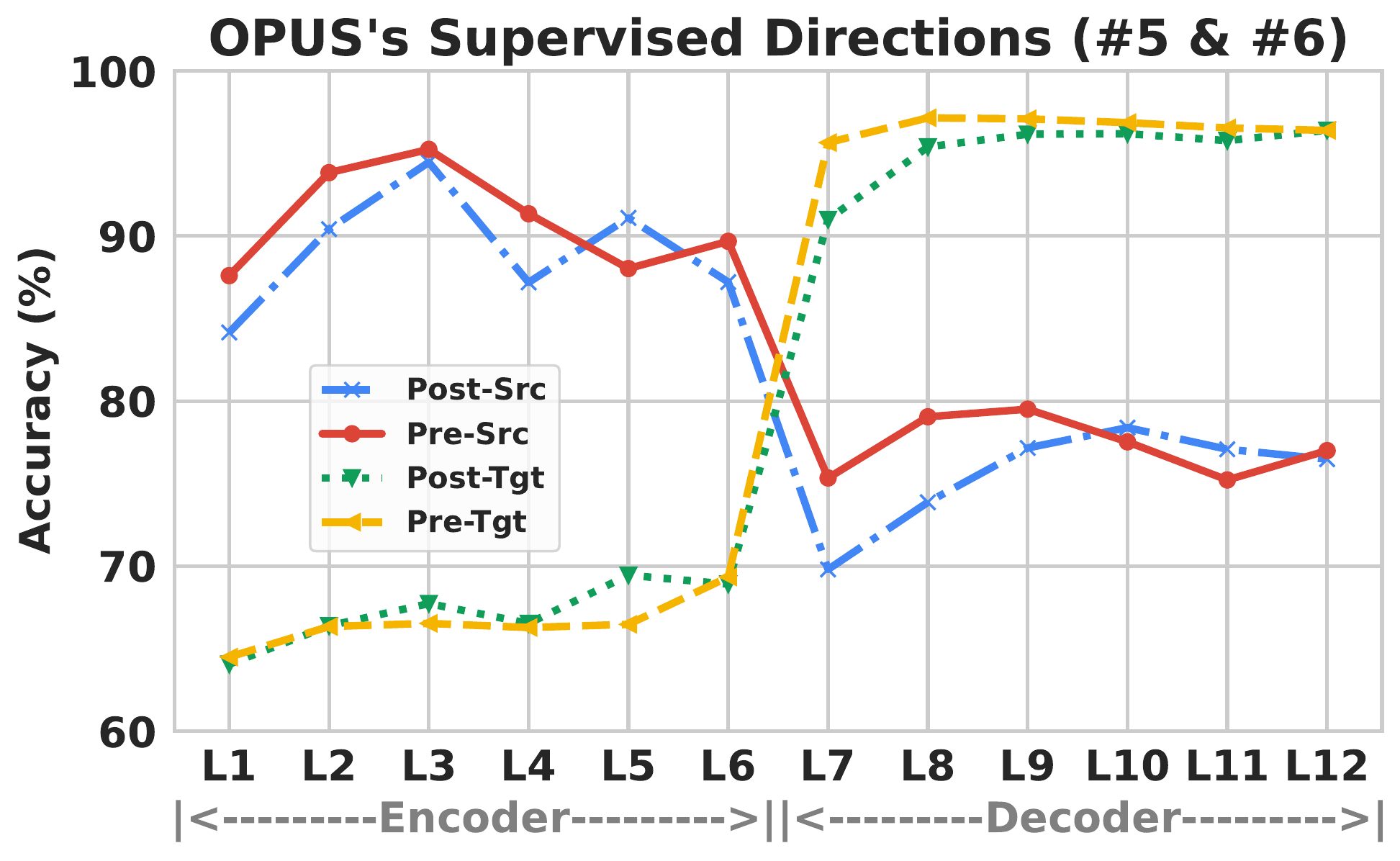}
\end{center}
\end{minipage}
\begin{minipage}[t]{0.33\textwidth}
\begin{center}
\includegraphics[width=\linewidth]{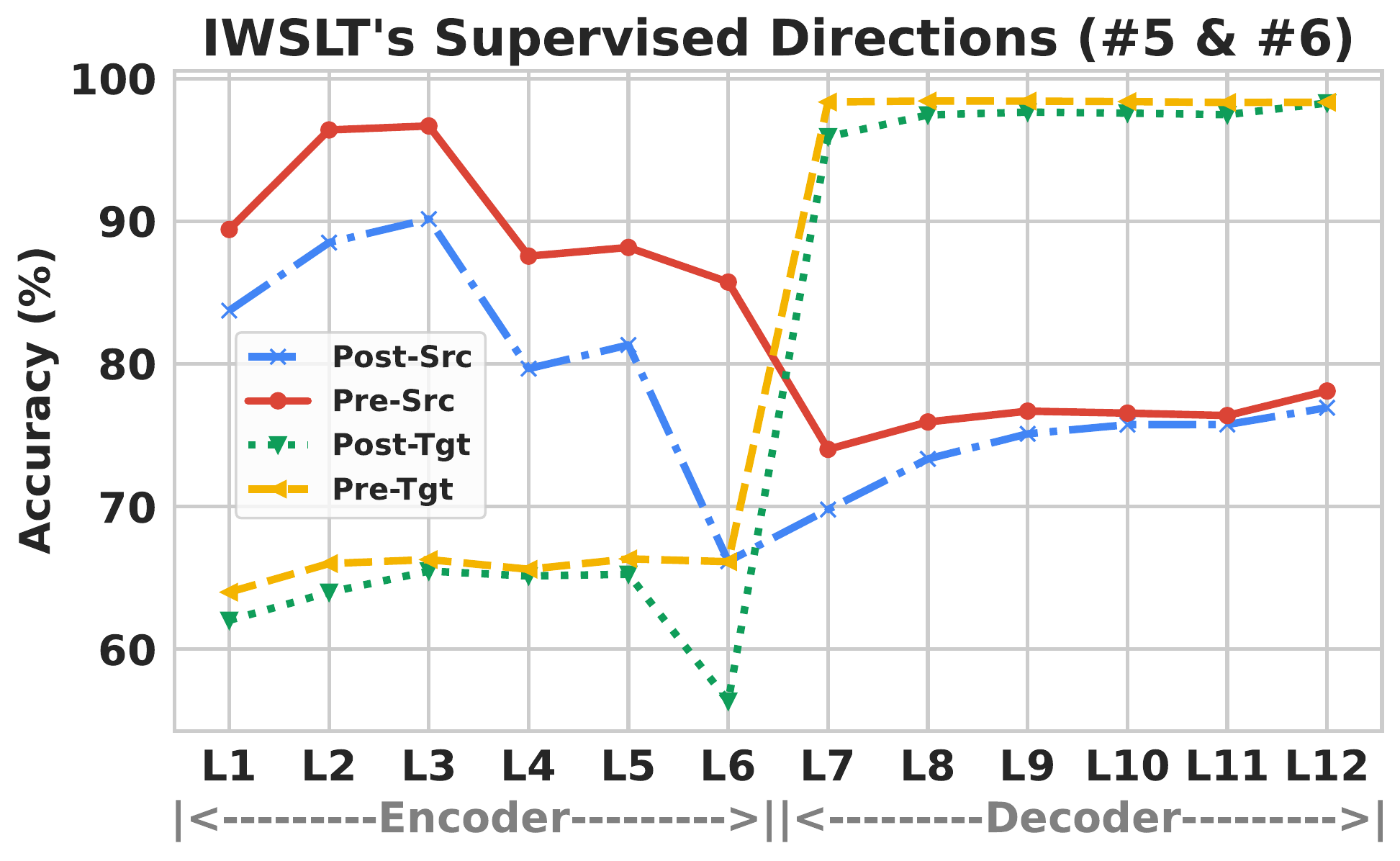}
\end{center}
\end{minipage}
\begin{minipage}[t]{0.33\textwidth}
\begin{center}
\includegraphics[width=\linewidth]{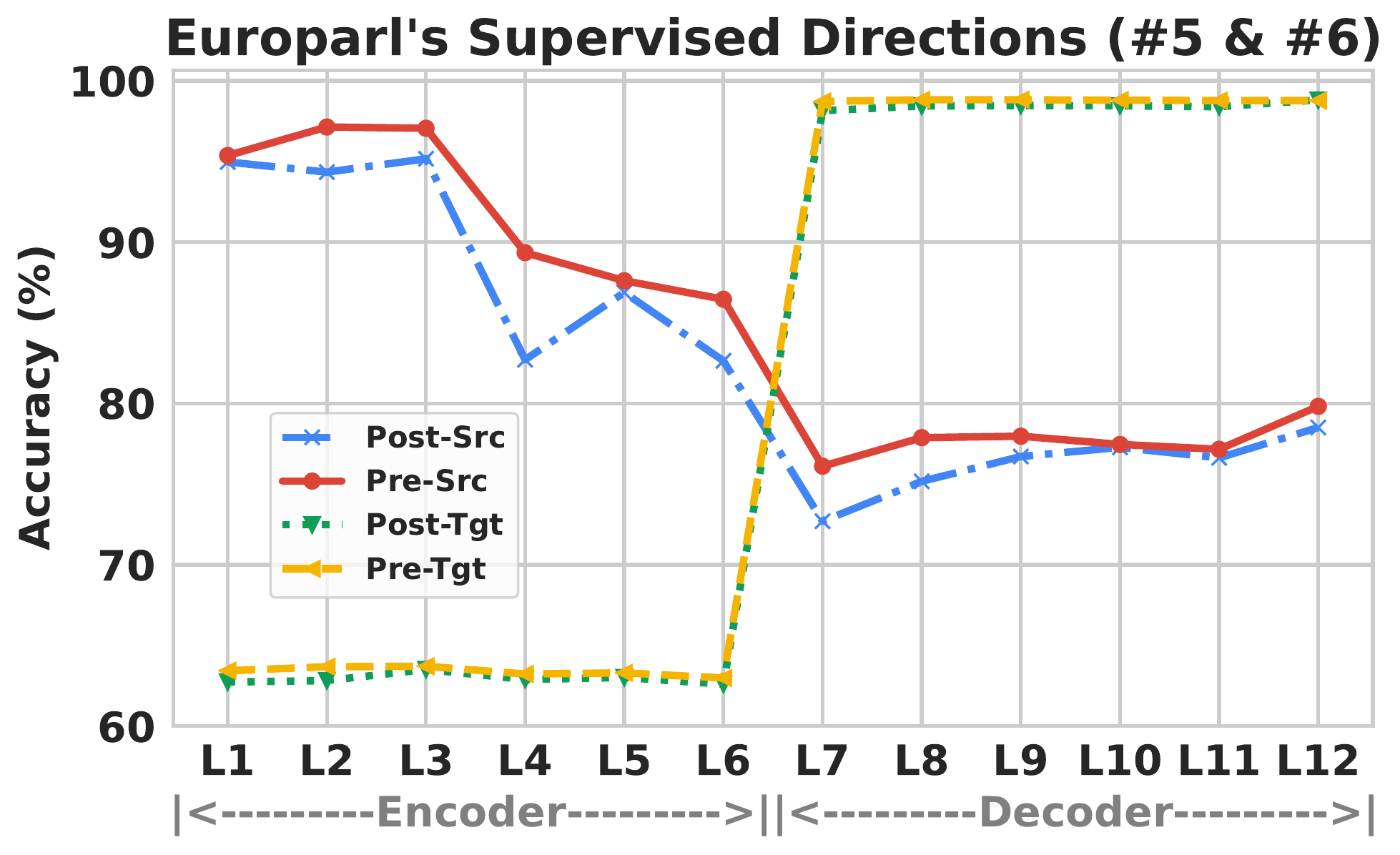}
\end{center}
\end{minipage}
\\
\begin{minipage}[t]{0.33\textwidth}
\begin{center}
\includegraphics[width=\linewidth]{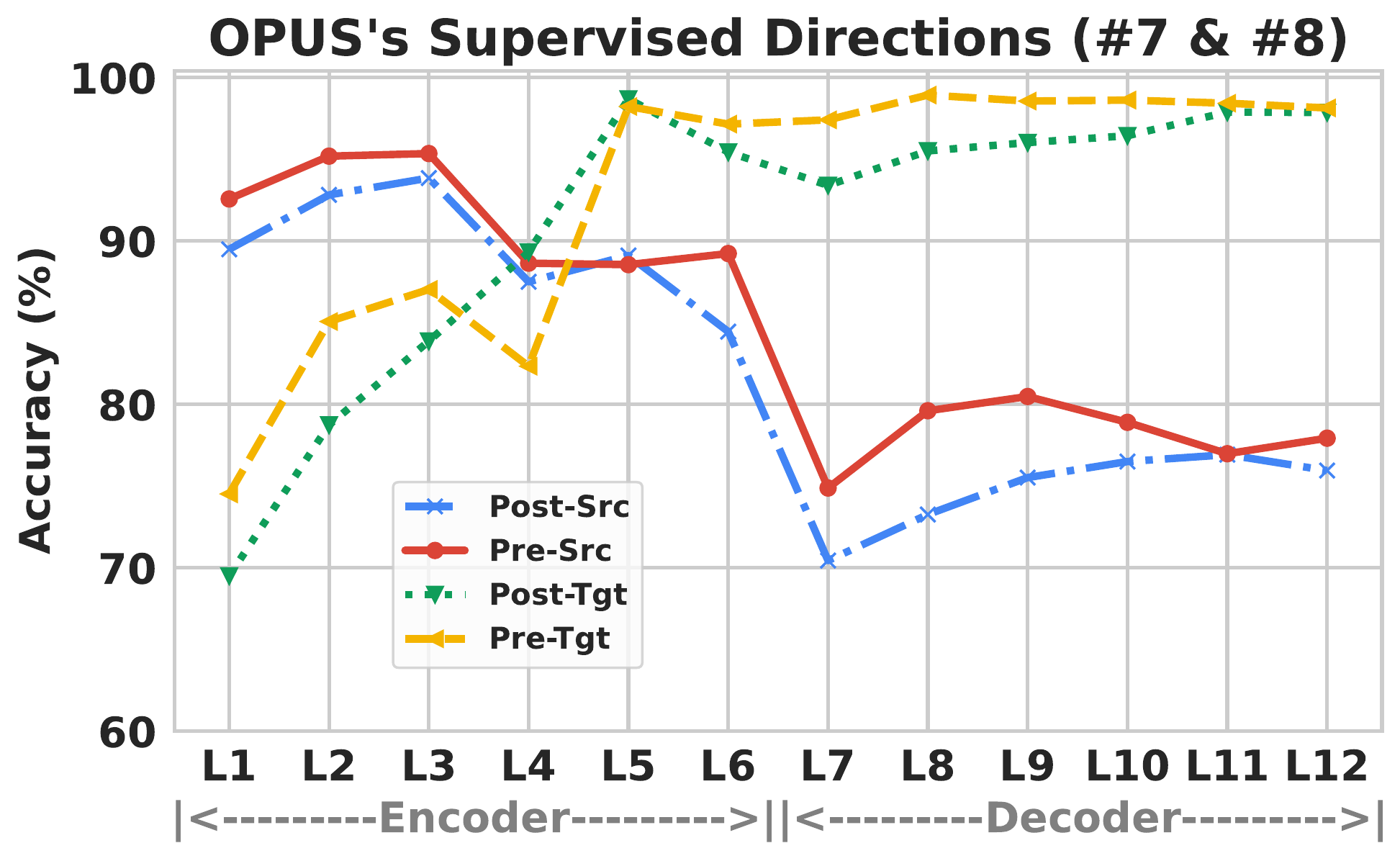}
\end{center}
\end{minipage}
\begin{minipage}[t]{0.33\textwidth}
\begin{center}
\includegraphics[width=\linewidth]{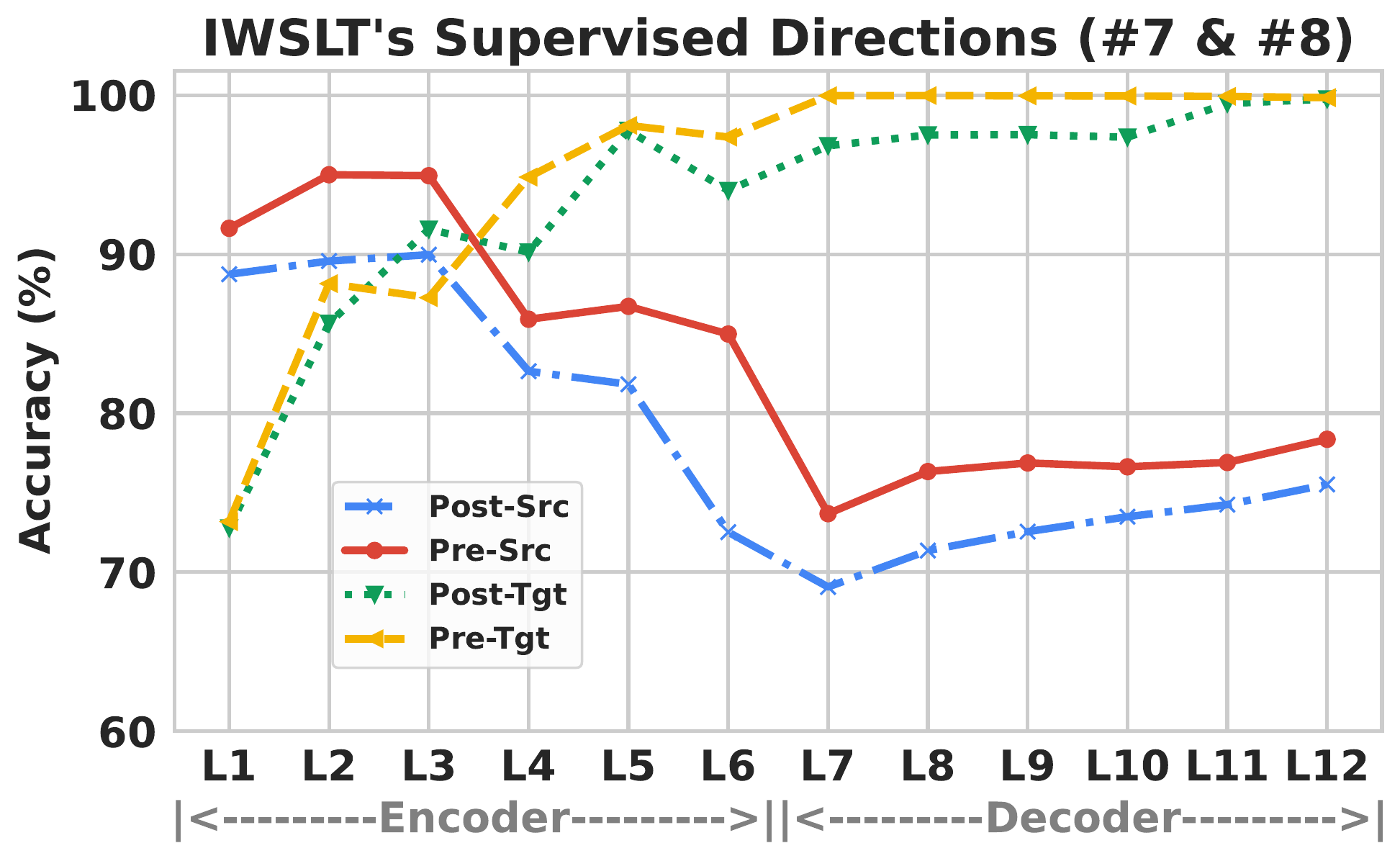}
\end{center}
\end{minipage}
\begin{minipage}[t]{0.33\textwidth}
\begin{center}
\includegraphics[width=\linewidth]{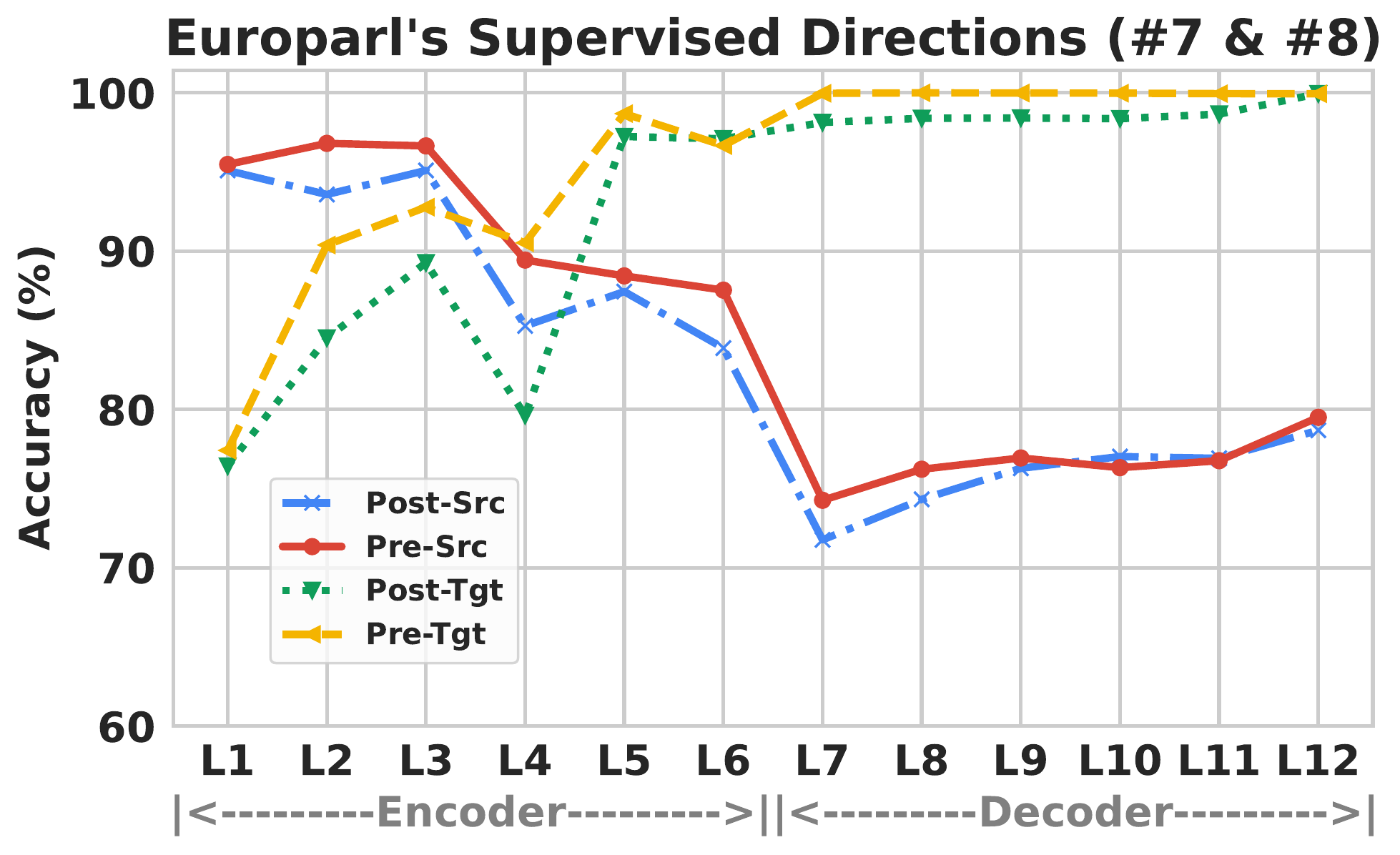}
\end{center}
\end{minipage}
\\
\begin{minipage}[t]{0.33\textwidth}
\begin{center}
\includegraphics[width=\linewidth]{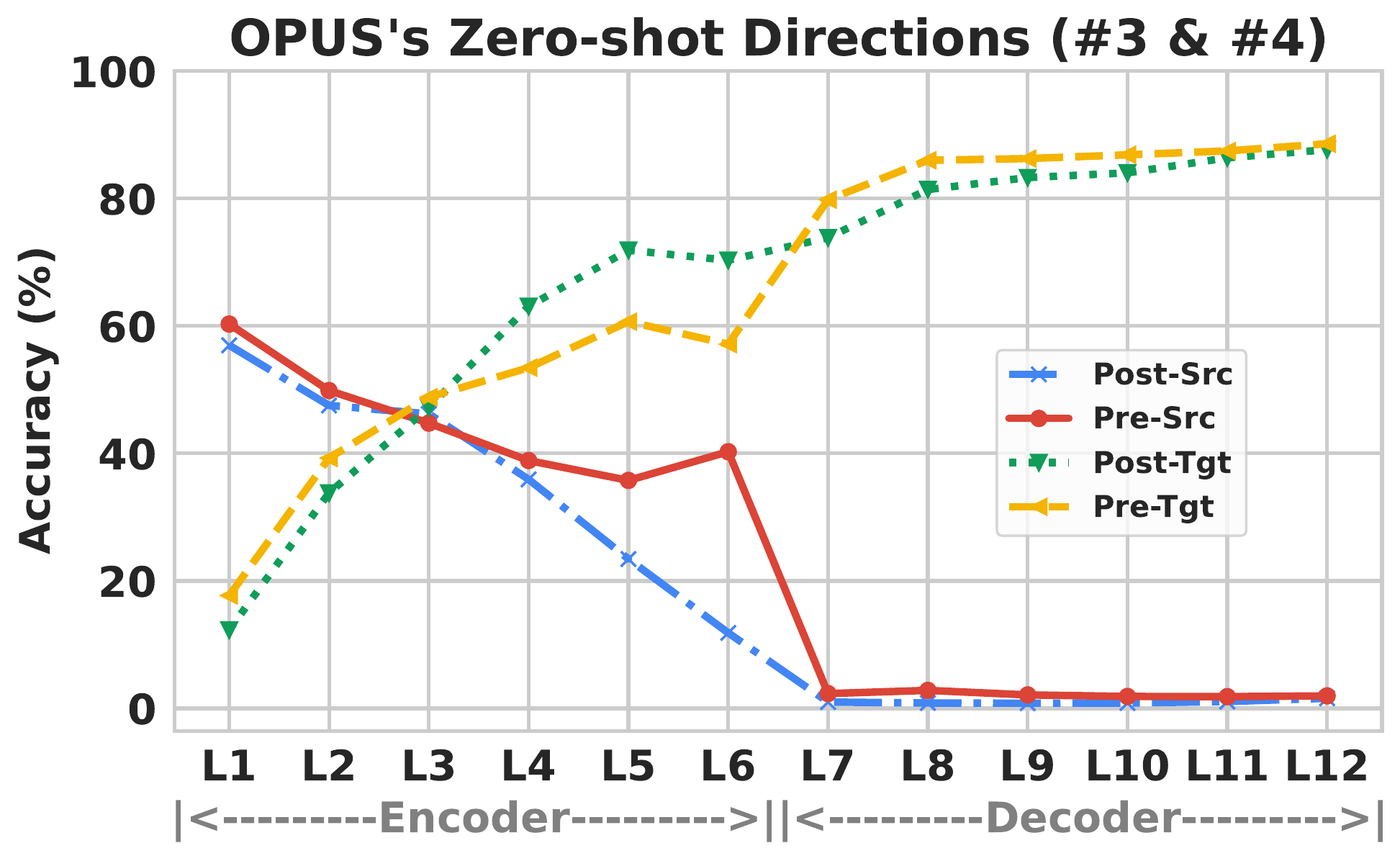}
\end{center}
\end{minipage}
\begin{minipage}[t]{0.33\textwidth}
\begin{center}
\includegraphics[width=\linewidth]{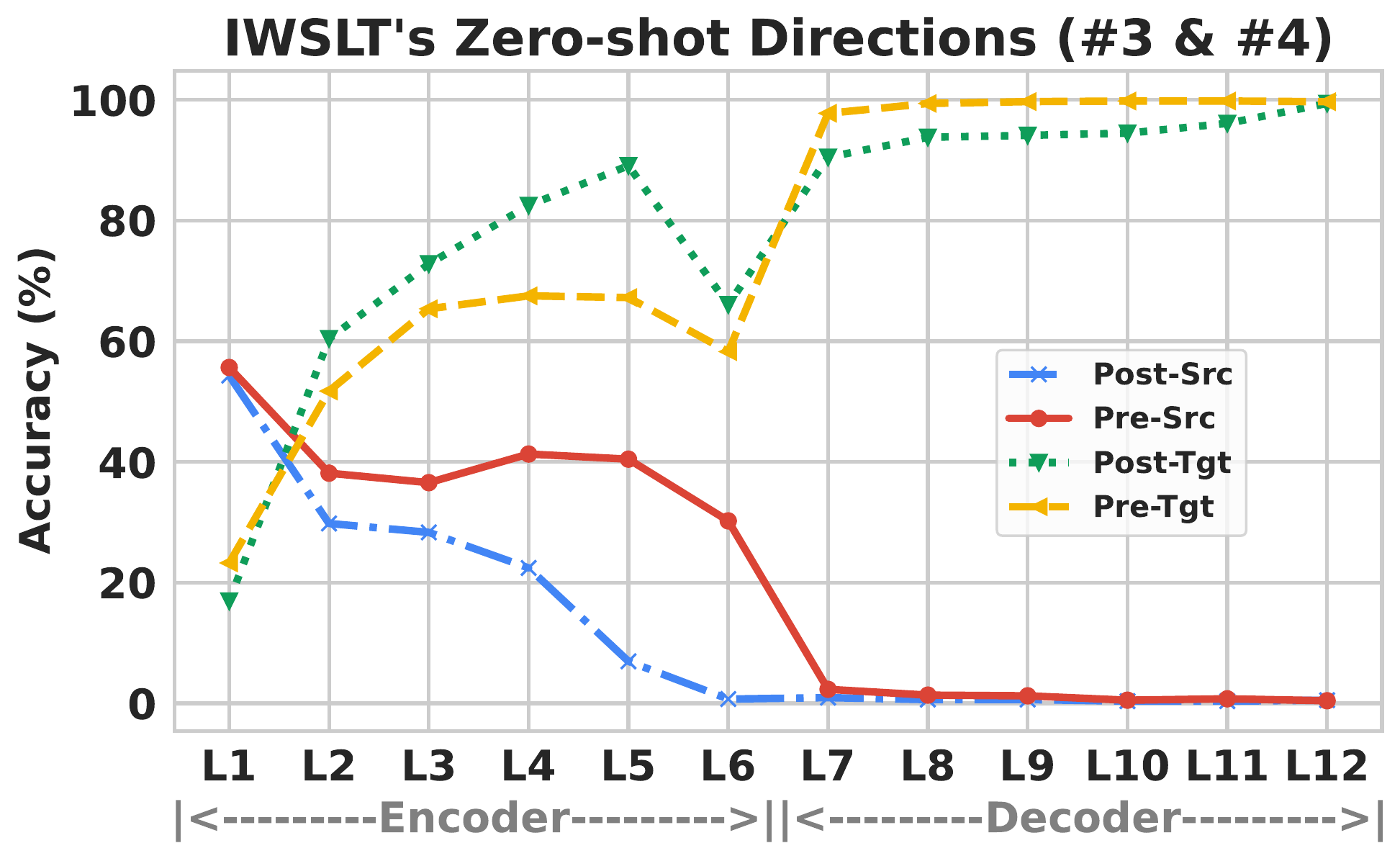}
\end{center}
\end{minipage}
\begin{minipage}[t]{0.33\textwidth}
\begin{center}
\includegraphics[width=\linewidth]{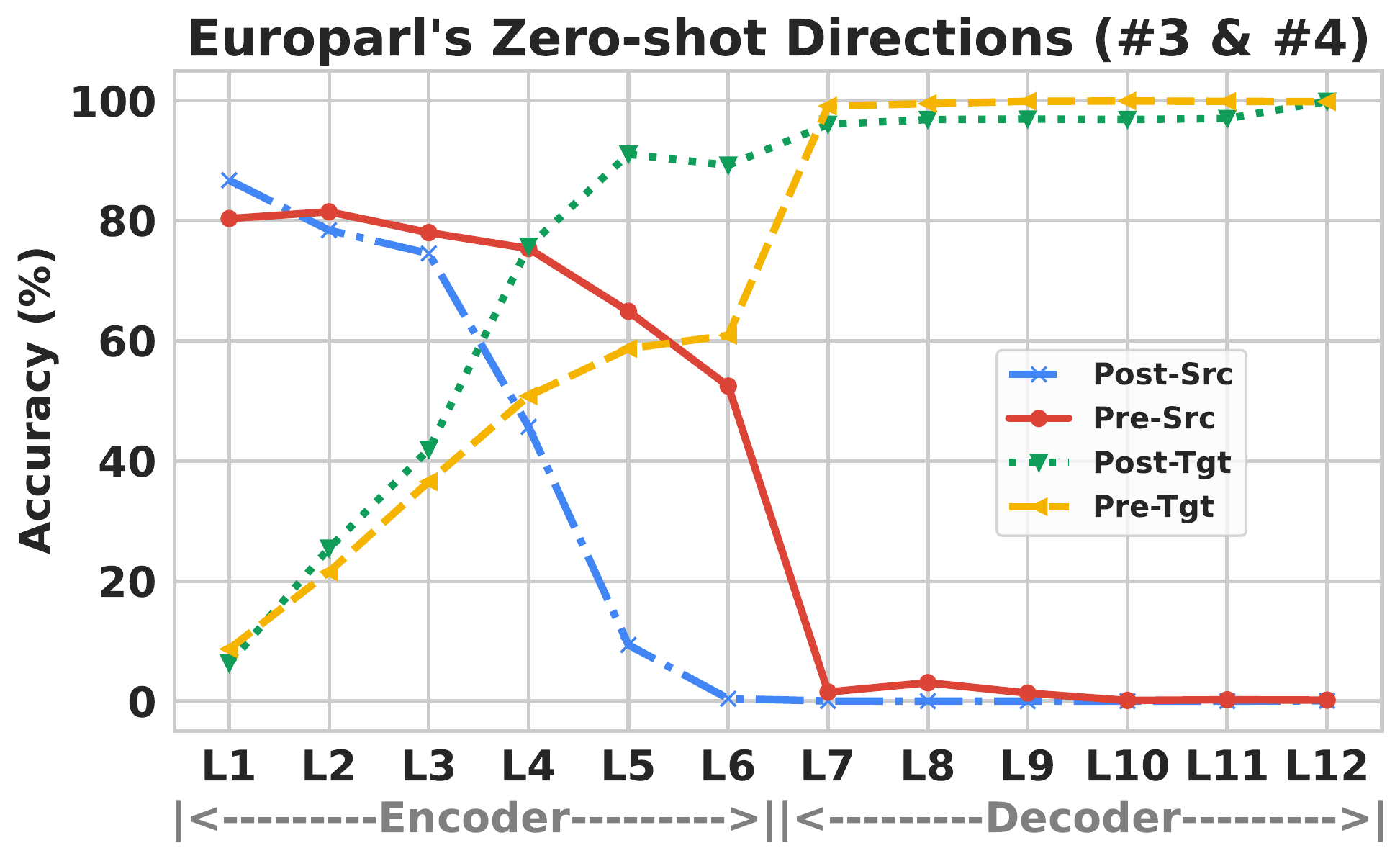}
\end{center}
\end{minipage}
\\
\begin{minipage}[t]{0.33\textwidth}
\begin{center}
\includegraphics[width=\linewidth]{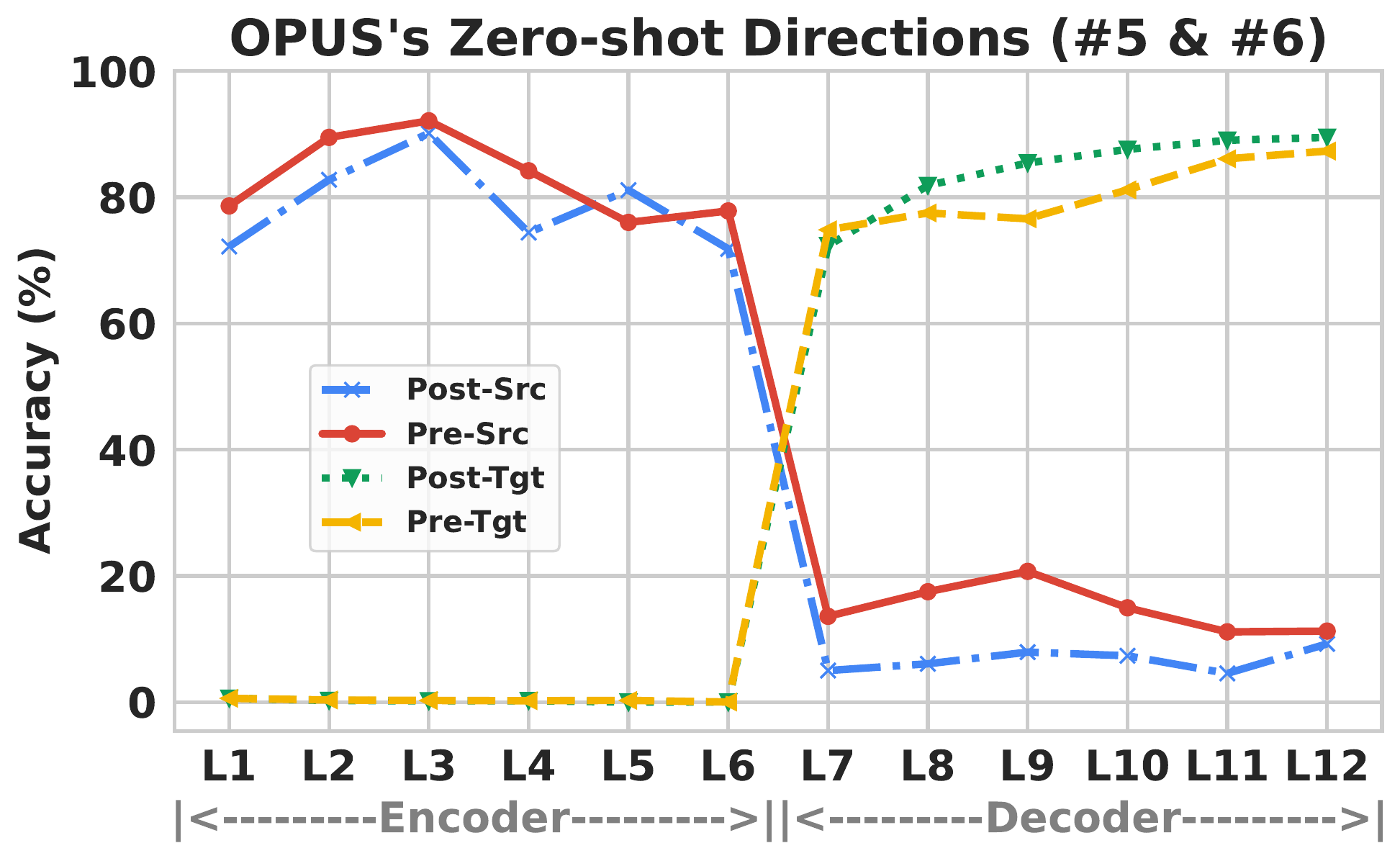}
\end{center}
\end{minipage}
\begin{minipage}[t]{0.33\textwidth}
\begin{center}
\includegraphics[width=\linewidth]{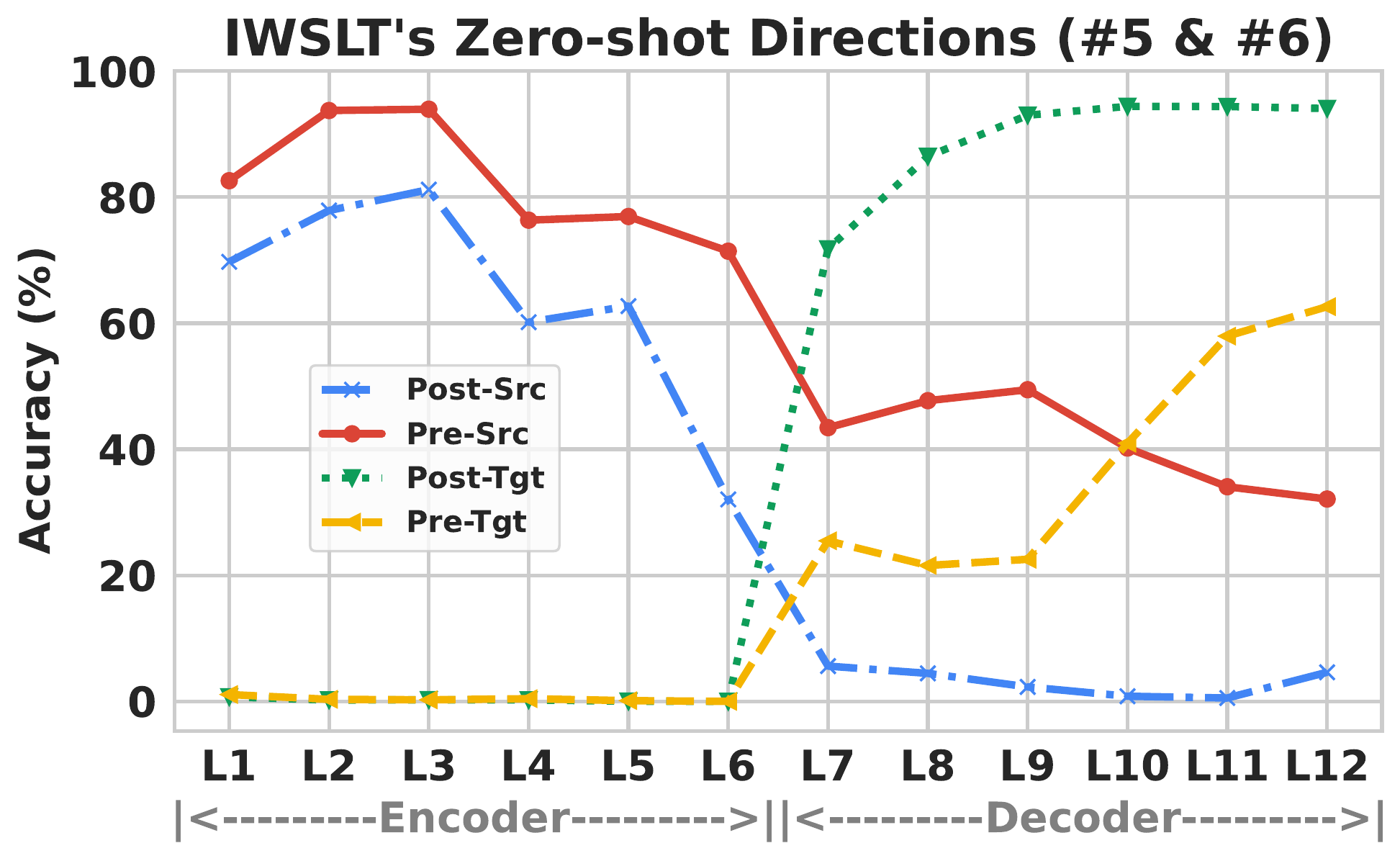}
\end{center}
\end{minipage}
\begin{minipage}[t]{0.33\textwidth}
\begin{center}
\includegraphics[width=\linewidth]{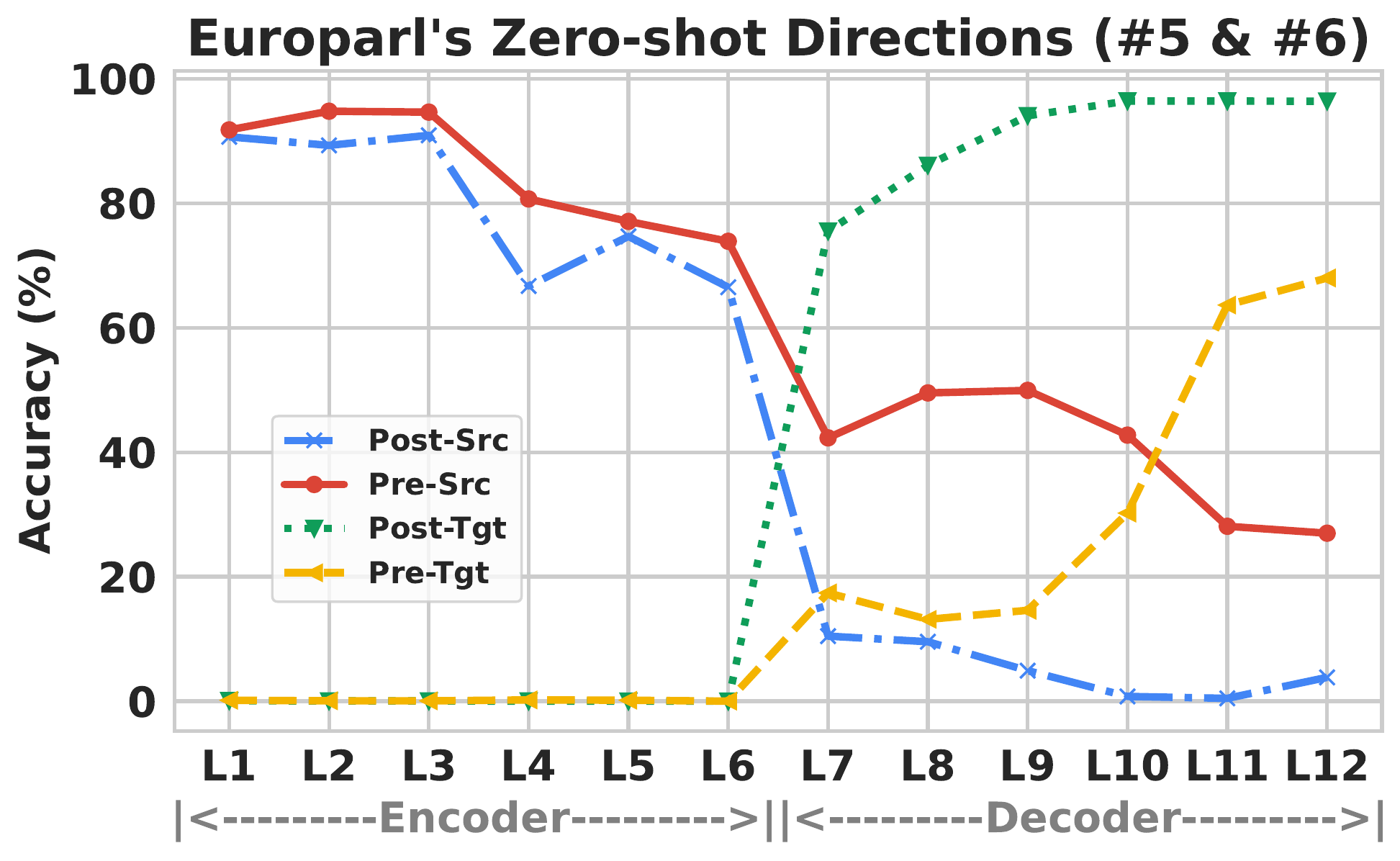}
\end{center}
\end{minipage}
\\
\begin{minipage}[t]{0.33\textwidth}
\begin{center}
\includegraphics[width=\linewidth]{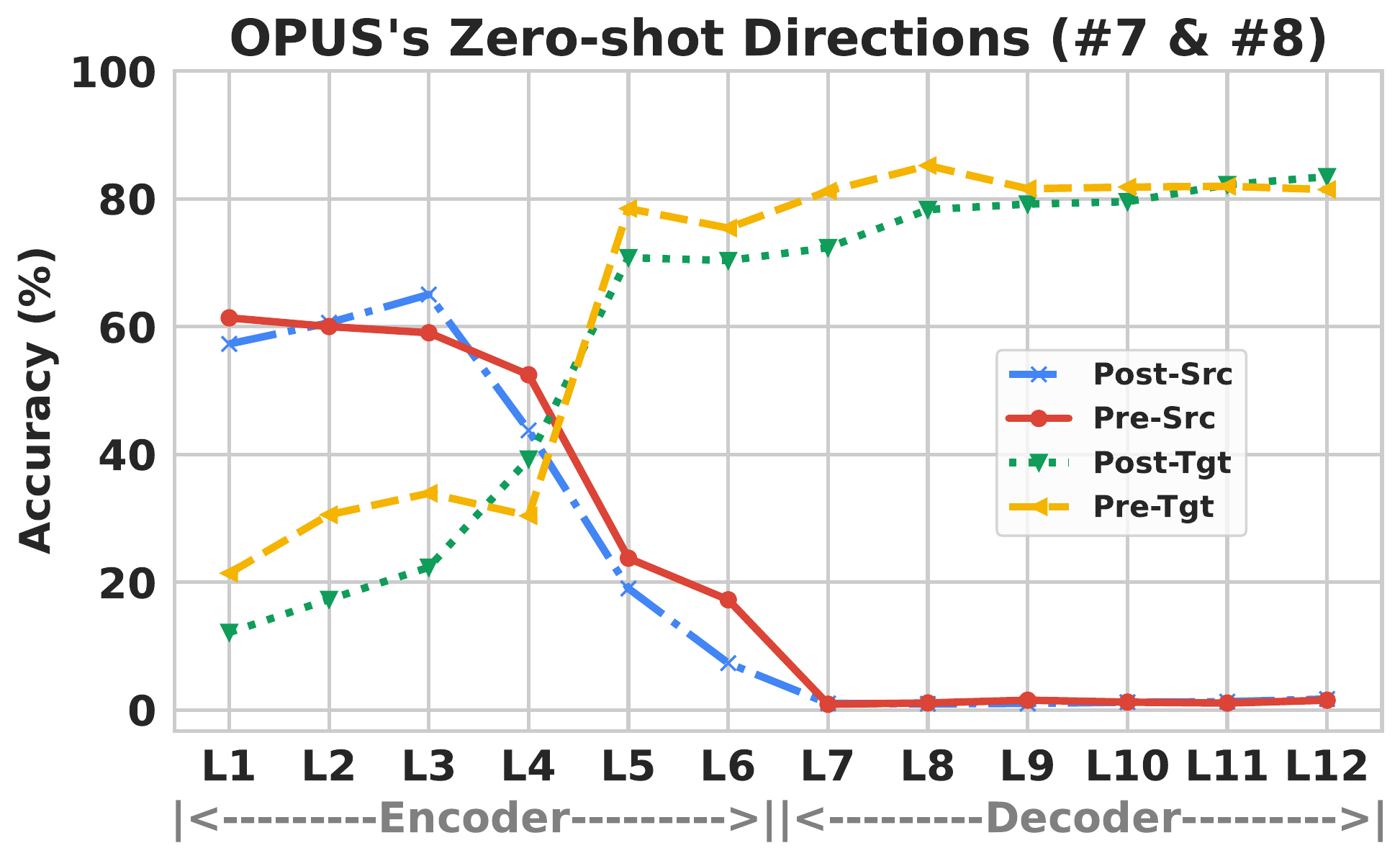}
\end{center}
\end{minipage}
\begin{minipage}[t]{0.33\textwidth}
\begin{center}
\includegraphics[width=\linewidth]{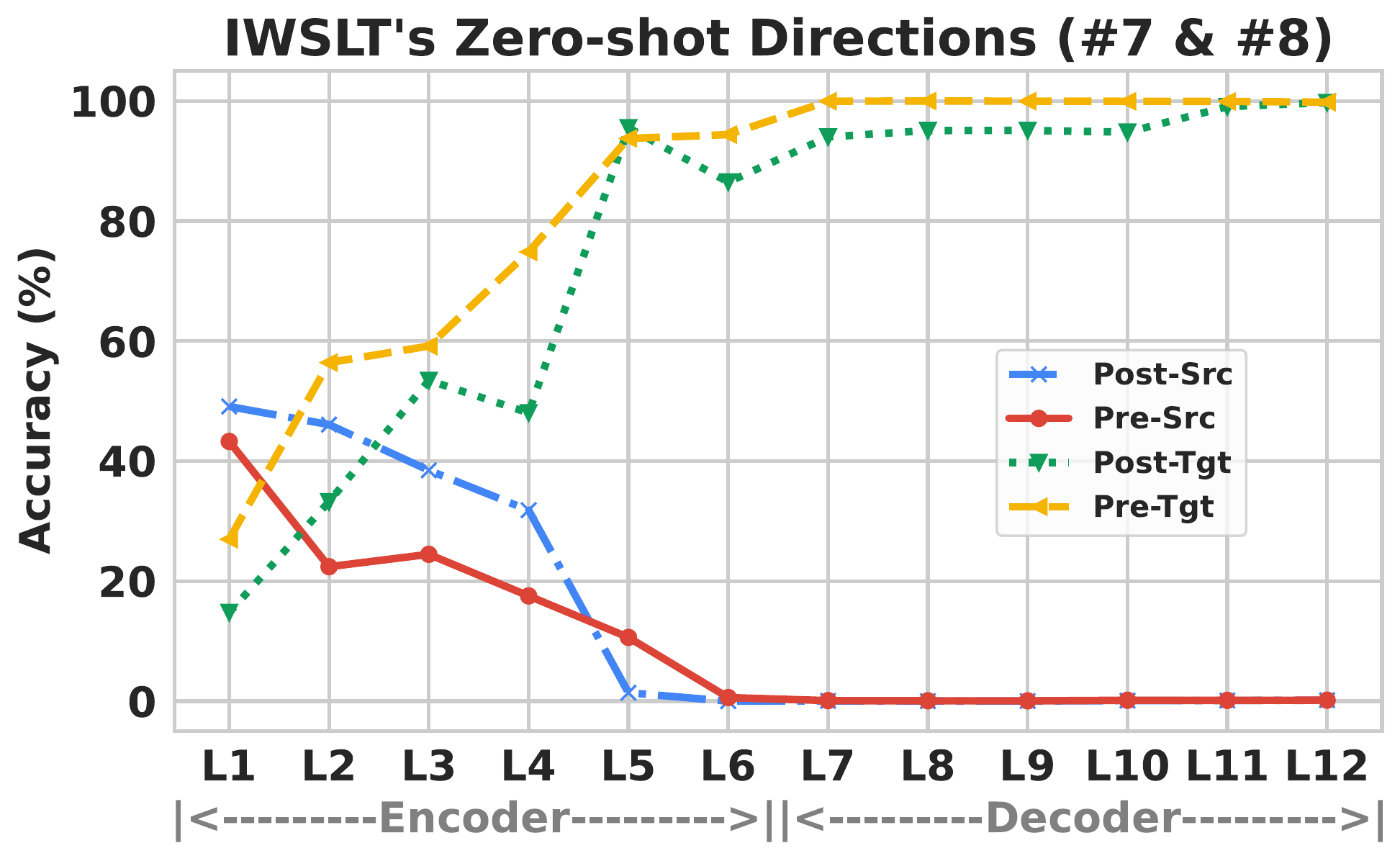}
\end{center}
\end{minipage}
\begin{minipage}[t]{0.33\textwidth}
\begin{center}
\includegraphics[width=\linewidth]{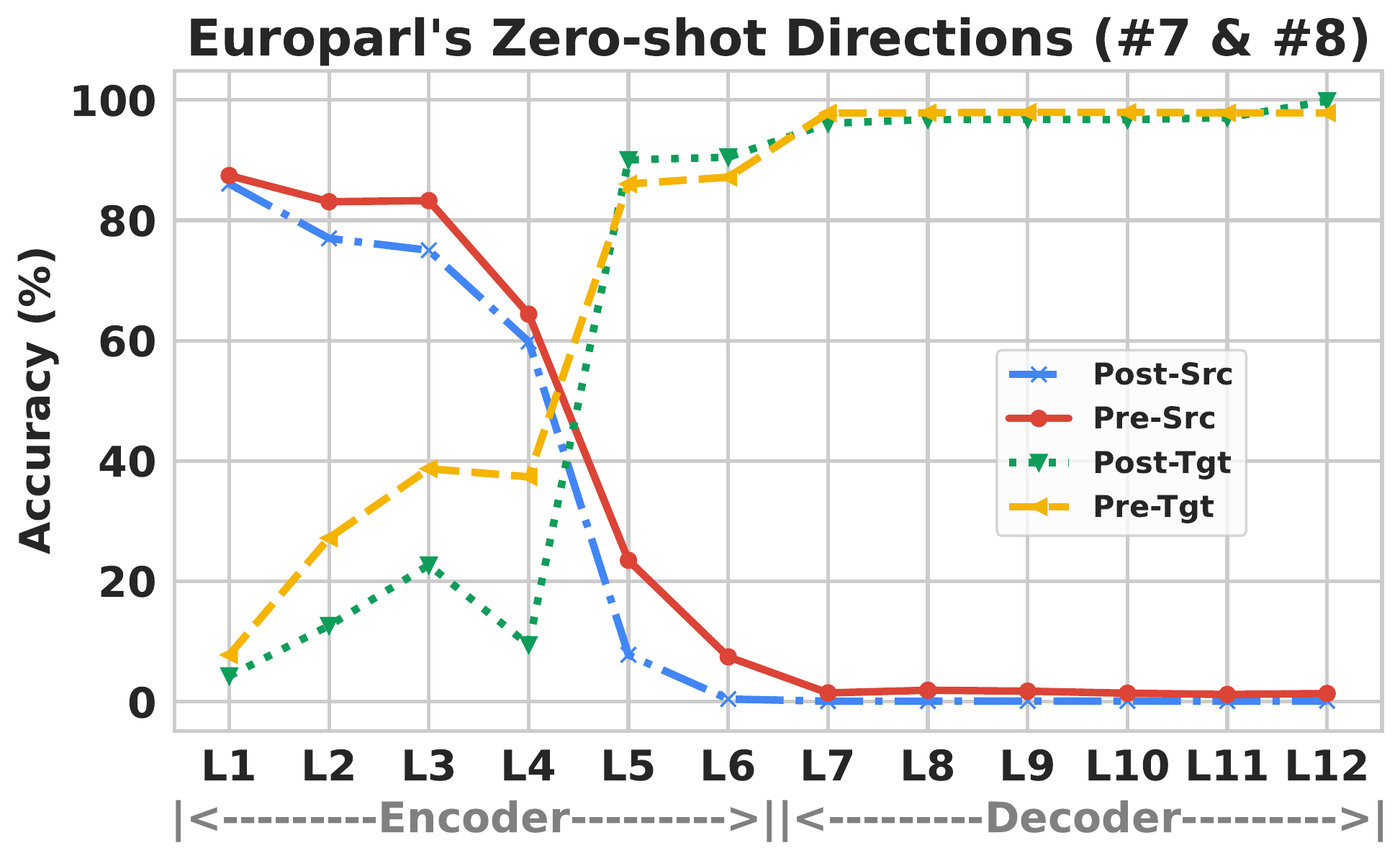}
\end{center}
\end{minipage}
\caption{\textbf{The LLR results of \#3 - \#8 (Table~\ref{tab:bleu1}) for both ZST and supervised directions for each dataset}. We report the average accuracy of three seeds and all the supervised or zero-shot directions. ``Pre-Src'' and ``Pre-Tgt'' indicate the layer-wise source and target language recognition for a PreNorm system (\#3, \#5, or \#7), while ``Post-Src'' and ``Post-Tgt'' denote similary for a PostNorm system (\#4, \#6, or \#8). ``L1'' to ``L6'' are $6$ encoder layers and ``L7'' to ``L12'' are $6$ decoder layers.}
\label{fig:off-target2}
\end{figure*}

We show the LLR results of \#3 - \#8 (Table~\ref{tab:bleu1}) for ZST and supervised directions in Fig.~\ref{fig:off-target2}.

\section{Details of the Main Results}
\label{app:res}
We report the specific BLEU score for each translation direction and each random seed in Tables~\ref{tab:bleu4},~\ref{tab:bleu5},~\ref{tab:bleu6},~\ref{tab:bleu7},~\ref{tab:bleu8}, and~\ref{tab:bleu9}.\footnote{Refer to details of setting random seeds in PyTorch at \url{https://pytorch.org/docs/stable/notes/randomness.html}.} In addition to BLEU scores, we present model-based evaluation results obtained using BLEURT~\cite{sellam-etal-2020-bleurt}\footnote{\url{https://github.com/google-research/bleurt}} in Table~\ref{tab:bleurt}. The results trend is consistent with those obtained from BLEU scores.

\begin{table*}[t]
    \centering
    \resizebox{0.97\linewidth}{!}{
    \begin{tabular}{l|l|rrrr|rrrr|rrrr|rrrr}
        \toprule
        Layer & \multirow{2}{*}{Direction} & \multicolumn{4}{c|}{S-ENC-T-DEC w/ Res.} & \multicolumn{4}{c|}{T-ENC w/ Res.} & \multicolumn{4}{c|}{S-ENC-T-DEC w/o Res.} & \multicolumn{4}{c}{T-ENC w/o Res.} \\
        Norm & & 1 & 10 & 20 & avg. & 1 & 10 & 20 & avg. & 1 & 10 & 20 & avg. & 1 & 10 & 20 & avg. \\
        \toprule
        \multirow{31}{*}{Pre.} & ar-de & 5.3 & 5.9 & 5.2 & 5.5 & 10.0 & 11.0 & 10.0 & 10.3 & 9.8 & 8.4 & 9.5 & 9.2 & 11.4 & 8.3 & 10.8 & 10.2 \\
        & ar-fr & 17.5 & 16.1 & 17.2 & 16.9 & 16.3 & 19.9 & 18.3 & 18.2 & 19.9 & 20.3 & 20.8 & 20.3 & 20.9 & 18.6 & 20.4 & 20.0 \\
        & ar-nl & 8.6 & 6.3 & 7.9 & 7.6 & 13.2 & 13.1 & 12.6 & 13.0 & 13.3 & 14.0 & 12.2 & 13.2 & 13.5 & 12.8 & 13.6 & 13.3 \\
        & ar-ru & 8.1 & 9.1 & 9.5 & 8.9 & 14.8 & 16.2 & 15.9 & 15.6 & 13.0 & 10.9 & 13.0 & 12.3 & 19.6 & 17.8 & 19.6 & 19.0 \\
        & ar-zh & 12.7 & 13.4 & 13.8 & 13.3 & 28.1 & 28.1 & 27.3 & 27.8 & 25.1 & 19.8 & 24.4 & 23.1 & 31.2 & 31.0 & 31.3 & 31.2 \\
        & de-ar & 3.6 & 3.3 & 2.5 & 3.1 & 5.6 & 5.0 & 3.9 & 4.8 & 6.9 & 6.4 & 5.6 & 6.3 & 6.4 & 5.1 & 3.3 & 4.9 \\
        & de-fr & 15.5 & 16.0 & 16.2 & 15.9 & 5.1 & 5.7 & 3.8 & 4.9 & 18.8 & 17.6 & 18.9 & 18.4 & 5.9 & 4.7 & 5.2 & 5.3 \\
        & de-nl & 19.4 & 15.9 & 18.8 & 18.0 & 12.4 & 8.9 & 8.6 & 10.0 & 21.4 & 20.4 & 20.7 & 20.8 & 9.1 & 7.1 & 7.7 & 8.0 \\
        & de-ru & 6.0 & 6.1 & 5.8 & 6.0 & 5.0 & 5.6 & 3.7 & 4.8 & 9.4 & 9.2 & 9.0 & 9.2 & 6.4 & 4.5 & 3.8 & 4.9 \\
        & de-zh & 7.6 & 9.5 & 8.8 & 8.6 & 15.6 & 12.4 & 11.9 & 13.3 & 14.4 & 12.8 & 13.2 & 13.5 & 16.4 & 4.1 & 6.0 & 8.8 \\
        & fr-ar & 9.5 & 7.5 & 8.5 & 8.5 & 15.5 & 16.2 & 13.2 & 15.0 & 15.4 & 13.1 & 14.4 & 14.3 & 18.5 & 16.5 & 15.8 & 16.9 \\
        & fr-de & 10.4 & 10.6 & 11.6 & 10.9 & 6.3 & 7.2 & 4.9 & 6.1 & 14.1 & 11.0 & 15.2 & 13.4 & 4.5 & 4.6 & 4.0 & 4.4 \\
        & fr-nl & 17.5 & 13.7 & 18.0 & 16.4 & 16.0 & 12.5 & 13.2 & 13.9 & 20.5 & 19.9 & 20.2 & 20.2 & 11.1 & 9.1 & 8.6 & 9.6 \\
        & fr-ru & 8.8 & 8.4 & 9.3 & 8.8 & 12.1 & 12.8 & 10.9 & 11.9 & 13.3 & 9.2 & 12.0 & 11.5 & 16.5 & 7.4 & 8.4 & 10.8 \\
        & fr-zh & 14.3 & 13.1 & 15.3 & 14.2 & 31.2 & 30.0 & 28.0 & 29.7 & 27.9 & 21.0 & 25.8 & 24.9 & 34.1 & 16.0 & 27.9 & 26.0 \\
        & nl-ar & 2.6 & 2.0 & 1.7 & 2.1 & 5.2 & 5.6 & 5.3 & 5.4 & 4.3 & 5.8 & 4.2 & 4.8 & 5.5 & 5.0 & 5.0 & 5.2 \\
        & nl-de & 14.3 & 14.4 & 13.9 & 14.2 & 12.8 & 13.9 & 11.3 & 12.7 & 16.9 & 14.8 & 18.3 & 16.7 & 13.8 & 6.9 & 10.9 & 10.5 \\
        & nl-fr & 18.3 & 17.4 & 18.5 & 18.1 & 13.1 & 16.1 & 12.4 & 13.9 & 21.5 & 19.9 & 22.3 & 21.2 & 15.0 & 7.1 & 13.8 & 12.0 \\
        & nl-ru & 4.2 & 4.4 & 3.4 & 4.0 & 9.5 & 9.8 & 8.6 & 9.3 & 7.2 & 6.5 & 7.3 & 7.0 & 10.3 & 6.6 & 7.3 & 8.1 \\
        & nl-zh & 2.2 & 3.2 & 3.0 & 2.8 & 10.8 & 10.0 & 10.4 & 10.4 & 7.0 & 8.0 & 6.3 & 7.1 & 11.1 & 7.5 & 10.0 & 9.5 \\
        & ru-ar & 9.7 & 7.6 & 7.6 & 8.3 & 15.6 & 16.1 & 14.6 & 15.4 & 15.9 & 13.3 & 14.0 & 14.4 & 18.6 & 19.1 & 18.0 & 18.6 \\
        & ru-de & 7.7 & 9.1 & 7.2 & 8.0 & 8.5 & 10.0 & 6.0 & 8.2 & 10.5 & 10.0 & 10.9 & 10.5 & 8.4 & 5.6 & 6.8 & 6.9 \\
        & ru-fr & 18.1 & 17.5 & 17.4 & 17.7 & 18.1 & 20.5 & 17.6 & 18.7 & 19.9 & 19.5 & 20.7 & 20.0 & 22.4 & 17.4 & 21.1 & 20.3 \\
        & ru-nl & 10.2 & 8.6 & 9.9 & 9.6 & 11.5 & 11.7 & 9.5 & 10.9 & 13.0 & 13.1 & 12.4 & 12.8 & 12.7 & 8.2 & 10.1 & 10.3 \\
        & ru-zh & 11.3 & 11.6 & 12.5 & 11.8 & 28.4 & 28.3 & 27.6 & 28.1 & 25.3 & 17.7 & 21.6 & 21.5 & 31.9 & 20.0 & 30.7 & 27.5 \\
        & zh-ar & 9.1 & 7.6 & 7.2 & 8.0 & 15.2 & 16.6 & 14.5 & 15.4 & 15.6 & 12.7 & 15.1 & 14.5 & 18.4 & 18.8 & 18.7 & 18.6 \\
        & zh-fr & 16.7 & 15.6 & 16.4 & 16.2 & 20.1 & 21.4 & 18.4 & 20.0 & 20.9 & 19.3 & 20.6 & 20.3 & 23.5 & 23.3 & 23.7 & 23.5 \\
        & zh-de & 4.7 & 5.8 & 5.4 & 5.3 & 7.8 & 8.1 & 7.0 & 7.6 & 7.5 & 6.9 & 7.1 & 7.2 & 8.6 & 8.6 & 8.8 & 8.7 \\
        & zh-nl & 6.9 & 5.4 & 6.0 & 6.1 & 8.6 & 8.6 & 8.2 & 8.5 & 8.5 & 8.0 & 8.0 & 8.2 & 9.1 & 9.2 & 8.8 & 9.0 \\
        & zh-ru & 6.9 & 8.2 & 7.8 & 7.6 & 13.7 & 15.7 & 12.9 & 14.1 & 12.8 & 10.0 & 11.8 & 11.5 & 18.7 & 19.8 & 19.7 & 19.4 \\
        & avg. & 10.3 & 9.8 & 10.2 & \textbf{10.1} & 13.5 & 13.9 & 12.4 & \textbf{13.3} & 15.0 & 13.3 & 14.5 & \textbf{14.3} & 15.1 & 11.7 & 13.3 & \textbf{13.4} \\
        \hline
        \multirow{31}{*}{Post.} & ar-de & 11.4 & 11.0 & 10.3 & 10.9 & 10.1 & 10.4 & 9.9 & 10.1 & 10.1 & 11.9 & 9.9 & 10.6 & 11.0 & 11.0 & 10.0 & 10.7 \\
        & ar-fr & 20.7 & 23.2 & 20.3 & 21.4 & 16.2 & 18.7 & 19.3 & 18.1 & 20.7 & 24.0 & 19.2 & 21.3 & 20.4 & 21.8 & 15.9 & 19.4 \\
        & ar-nl & 13.3 & 13.7 & 12.5 & 13.2 & 12.8 & 13.5 & 13.3 & 13.2 & 13.4 & 14.4 & 12.5 & 13.4 & 13.2 & 13.9 & 13.0 & 13.4 \\
        & ar-ru & 16.9 & 18.7 & 16.1 & 17.2 & 17.4 & 17.2 & 18.6 & 17.7 & 13.5 & 19.1 & 14.7 & 15.8 & 20.4 & 20.7 & 18.7 & 19.9 \\
        & ar-zh & 28.6 & 29.4 & 29.2 & 29.1 & 29.2 & 30.4 & 30.3 & 30.0 & 26.1 & 30.7 & 27.4 & 28.1 & 32.9 & 32.9 & 31.9 & 32.6 \\
        & de-ar & 7.2 & 7.2 & 6.6 & 7.0 & 5.7 & 5.6 & 5.8 & 5.7 & 6.9 & 7.6 & 7.6 & 7.4 & 4.4 & 4.1 & 3.1 & 3.9 \\
        & de-fr & 17.6 & 19.3 & 18.2 & 18.4 & 5.1 & 6.6 & 5.8 & 5.8 & 17.3 & 20.3 & 17.3 & 18.3 & 5.4 & 7.9 & 4.1 & 5.8 \\
        & de-nl & 21.4 & 21.8 & 20.4 & 21.2 & 9.1 & 9.5 & 7.9 & 8.8 & 20.0 & 22.3 & 20.5 & 20.9 & 9.7 & 11.9 & 7.1 & 9.6 \\
        & de-ru & 12.3 & 13.8 & 12.8 & 13.0 & 6.0 & 6.3 & 7.2 & 6.5 & 10.1 & 13.3 & 10.5 & 11.3 & 5.2 & 4.0 & 3.7 & 4.3 \\
        & de-zh & 16.1 & 16.9 & 16.5 & 16.5 & 8.9 & 15.3 & 15.0 & 13.1 & 11.2 & 16.9 & 13.5 & 13.9 & 14.1 & 11.1 & 3.1 & 9.4 \\
        & fr-ar & 17.9 & 17.8 & 18.9 & 18.2 & 16.4 & 17.1 & 16.4 & 16.6 & 14.6 & 19.5 & 16.3 & 16.8 & 16.4 & 16.6 & 14.8 & 15.9 \\
        & fr-de & 15.0 & 17.3 & 17.0 & 16.4 & 5.4 & 6.7 & 6.5 & 6.2 & 13.1 & 17.0 & 13.5 & 14.5 & 4.9 & 7.0 & 4.8 & 5.6 \\
        & fr-nl & 21.4 & 21.8 & 20.3 & 21.2 & 11.3 & 13.3 & 11.6 & 12.1 & 20.6 & 22.7 & 20.5 & 21.3 & 11.6 & 14.1 & 10.1 & 11.9 \\
        & fr-ru & 17.7 & 19.5 & 15.9 & 17.7 & 16.7 & 13.3 & 18.5 & 16.2 & 12.9 & 20.7 & 13.3 & 15.6 & 10.9 & 15.5 & 13.3 & 13.2 \\
        & fr-zh & 30.5 & 32.0 & 31.8 & 31.4 & 29.8 & 32.0 & 31.4 & 31.1 & 25.9 & 32.5 & 28.4 & 28.9 & 31.7 & 32.0 & 30.3 & 31.3 \\
        & nl-ar & 5.3 & 5.9 & 5.6 & 5.6 & 6.0 & 5.3 & 5.8 & 5.7 & 5.2 & 6.1 & 6.4 & 5.9 & 5.0 & 5.2 & 4.5 & 4.9 \\
        & nl-de & 17.9 & 19.7 & 19.1 & 18.9 & 10.9 & 12.8 & 10.5 & 11.4 & 16.5 & 19.8 & 17.1 & 17.8 & 9.0 & 10.4 & 10.4 & 9.9 \\
        & nl-fr & 21.1 & 22.5 & 21.2 & 21.6 & 13.8 & 13.4 & 13.0 & 13.4 & 21.2 & 22.9 & 19.6 & 21.2 & 10.1 & 12.6 & 9.5 & 10.7 \\
        & nl-ru & 10.0 & 11.2 & 10.2 & 10.5 & 9.7 & 9.1 & 8.8 & 9.2 & 8.4 & 10.9 & 8.6 & 9.3 & 8.6 & 7.6 & 8.2 & 8.1 \\
        & nl-zh & 9.6 & 11.1 & 9.6 & 10.1 & 10.2 & 10.4 & 10.0 & 10.2 & 5.4 & 11.1 & 7.3 & 7.9 & 9.9 & 9.9 & 7.5 & 9.1 \\
        & ru-ar & 18.7 & 18.7 & 18.2 & 18.5 & 16.9 & 17.9 & 17.5 & 17.4 & 14.8 & 19.7 & 16.2 & 16.9 & 17.9 & 18.9 & 17.0 & 17.9 \\
        & ru-de & 12.9 & 12.9 & 12.9 & 12.9 & 8.7 & 8.1 & 9.0 & 8.6 & 10.8 & 13.3 & 10.5 & 11.5 & 8.6 & 9.2 & 7.9 & 8.6 \\
        & ru-fr & 21.5 & 24.0 & 21.2 & 22.2 & 19.4 & 17.9 & 19.0 & 18.8 & 20.1 & 24.8 & 19.0 & 21.3 & 16.8 & 22.0 & 13.8 & 17.5 \\
        & ru-nl & 13.0 & 13.6 & 12.7 & 13.1 & 10.9 & 11.8 & 12.4 & 11.7 & 13.3 & 14.2 & 13.0 & 13.5 & 11.0 & 12.0 & 9.7 & 10.9 \\
        & ru-zh & 27.6 & 29.8 & 28.6 & 28.7 & 30.1 & 30.4 & 30.6 & 30.4 & 23.6 & 30.2 & 24.6 & 26.1 & 32.5 & 32.2 & 29.0 & 31.2 \\
        & zh-ar & 18.0 & 17.4 & 17.3 & 17.6 & 16.9 & 17.5 & 17.1 & 17.2 & 16.3 & 19.3 & 17.0 & 17.5 & 19.1 & 19.8 & 19.4 & 19.4 \\
        & zh-fr & 20.2 & 21.3 & 20.2 & 20.6 & 21.4 & 22.3 & 21.5 & 21.7 & 20.5 & 24.1 & 18.3 & 21.0 & 23.1 & 24.4 & 24.5 & 24.0 \\
        & zh-de & 8.6 & 9.1 & 8.8 & 8.8 & 7.3 & 7.4 & 7.1 & 7.3 & 8.3 & 9.9 & 7.5 & 8.6 & 8.7 & 8.5 & 8.0 & 8.4 \\
        & zh-nl & 8.7 & 8.5 & 8.1 & 8.4 & 8.9 & 8.7 & 8.4 & 8.7 & 8.9 & 9.0 & 8.1 & 8.7 & 8.9 & 9.3 & 9.0 & 9.1 \\
        & zh-ru & 15.3 & 15.8 & 14.1 & 15.1 & 16.7 & 17.3 & 17.6 & 17.2 & 13.3 & 17.8 & 12.8 & 14.6 & 20.2 & 20.5 & 20.2 & 20.3 \\
        & avg. & 16.5 & 17.5 & 16.5 & \textbf{16.8} & 13.6 & 14.2 & 14.2 & \textbf{14.0} & 14.8 & 18.2 & 15.0 & \textbf{16.0} & 14.1 & 14.9 & 12.8 & \textbf{13.9} \\
        \bottomrule
    \end{tabular}
    }
    \caption{\textbf{BLEU scores of OPUS in ZST directions}. Scores in \textbf{bold} are the results reported in Table~\ref{tab:bleu1}. ``1,'' ``10,'' and ``20'' indicates three random seeds. ``Res.'' indicates the residual connection of self-attention in the $4^{th}$ encoder layer.}
    \label{tab:bleu4}
\end{table*}

\begin{table*}[t]
    \centering
    \resizebox{\linewidth}{!}{
    \begin{tabular}{l|l|rrrr|rrrr|rrrr|rrrr}
        \toprule
        Layer & \multirow{2}{*}{Direction} & \multicolumn{4}{c|}{S-ENC-T-DEC w/ Res.} & \multicolumn{4}{c|}{T-ENC w/ Res.} & \multicolumn{4}{c|}{S-ENC-T-DEC w/o Res.} & \multicolumn{4}{c}{T-ENC w/o Res.} \\
        Norm & & 1 & 10 & 20 & avg. & 1 & 10 & 20 & avg. & 1 & 10 & 20 & avg. & 1 & 10 & 20 & avg. \\
        \toprule
        \multirow{13}{*}{Pre.} & en-ar & 23.6 & 24.1 & 23.2 & 23.6 & 23.7 & 23.9 & 24.1 & 23.9 & 24.0 & 23.2 & 23.1 & 23.4 & 22.8 & 23.8 & 23.8 & 23.5 \\
        & ar-en & 37.6 & 37.1 & 37.3 & 37.3 & 37.5 & 37.1 & 37.5 & 37.4 & 37.4 & 37.2 & 36.9 & 37.2 & 36.4 & 36.7 & 37.0 & 36.7 \\
        & en-de & 29.7 & 30.1 & 30.4 & 30.1 & 30.4 & 29.6 & 30.4 & 30.1 & 30.1 & 30.1 & 30.1 & 30.1 & 30.3 & 30.5 & 30.7 & 30.5 \\
        & de-en & 34.3 & 34.5 & 34.2 & 34.3 & 34.5 & 34.1 & 34.3 & 34.3 & 35.0 & 34.7 & 34.3 & 34.7 & 33.8 & 34.1 & 34.4 & 34.1 \\
        & en-fr & 33.5 & 33.7 & 33.6 & 33.6 & 33.4 & 33.8 & 33.6 & 33.6 & 33.7 & 33.1 & 33.8 & 33.5 & 33.0 & 33.6 & 33.1 & 33.2 \\
        & fr-en & 35.6 & 35.4 & 35.3 & 35.4 & 35.0 & 35.0 & 35.5 & 35.2 & 35.6 & 35.2 & 35.1 & 35.3 & 34.4 & 35.2 & 35.0 & 34.9 \\
        & en-nl & 27.7 & 28.4 & 28.2 & 28.1 & 28.4 & 27.9 & 28.3 & 28.2 & 27.6 & 28.0 & 27.9 & 27.8 & 28.1 & 28.1 & 28.0 & 28.1 \\
        & nl-en & 31.3 & 30.8 & 31.2 & 31.1 & 30.9 & 30.7 & 30.8 & 30.8 & 31.0 & 30.8 & 31.0 & 30.9 & 30.4 & 30.9 & 30.5 & 30.6 \\
        & en-ru & 29.2 & 29.7 & 29.6 & 29.5 & 29.4 & 29.8 & 29.8 & 29.7 & 29.5 & 29.1 & 29.6 & 29.4 & 29.4 & 29.9 & 29.2 & 29.5 \\
        & ru-en & 35.2 & 34.6 & 35.0 & 34.9 & 34.7 & 34.6 & 35.0 & 34.8 & 35.2 & 34.8 & 35.1 & 35.0 & 34.3 & 34.8 & 34.7 & 34.6 \\
        & en-zh & 40.7 & 40.8 & 40.9 & 40.8 & 40.6 & 40.3 & 40.7 & 40.5 & 40.7 & 40.4 & 40.6 & 40.6 & 39.6 & 40.7 & 40.6 & 40.3 \\
        & zh-en & 46.2 & 46.1 & 45.9 & 46.1 & 46.1 & 46.1 & 46.2 & 46.1 & 46.2 & 45.9 & 45.8 & 46.0 & 45.6 & 46.4 & 46.3 & 46.1 \\
        & avg. & 33.7 & 33.8 & 33.7 & \textbf{33.7} & 33.7 & 33.6 & 33.9 & \textbf{33.7} & 33.8 & 33.5 & 33.6 & \textbf{33.7} & 33.2 & 33.7 & 33.6 & \textbf{33.5} \\
        \hline
        \multirow{13}{*}{Post.} & en-ar & 23.9 & 23.4 & 23.7 & 23.7 & 24.6 & 24.4 & 24.3 & 24.4 & 23.7 & 23.8 & 23.8 & 23.8 & 24.0 & 23.8 & 24.0 & 23.9 \\
        & ar-en & 37.8 & 37.3 & 37.5 & 37.5 & 37.8 & 37.5 & 37.2 & 37.5 & 37.7 & 37.2 & 37.6 & 37.5 & 37.8 & 37.3 & 37.7 & 37.6 \\
        & en-de & 30.8 & 31.0 & 29.3 & 30.4 & 31.2 & 29.9 & 31.2 & 30.8 & 31.1 & 30.5 & 31.2 & 30.9 & 31.1 & 30.5 & 31.5 & 31.0 \\
        & de-en & 34.6 & 34.6 & 34.8 & 34.7 & 34.9 & 34.6 & 34.7 & 34.7 & 34.8 & 34.6 & 34.7 & 34.7 & 34.4 & 34.6 & 34.4 & 34.5 \\
        & en-fr & 33.9 & 33.4 & 34.1 & 33.8 & 34.1 & 33.8 & 33.9 & 33.9 & 33.5 & 33.5 & 33.2 & 33.4 & 33.7 & 33.8 & 33.6 & 33.7 \\
        & fr-en & 35.5 & 35.6 & 35.4 & 35.5 & 35.6 & 35.7 & 35.4 & 35.6 & 35.0 & 35.5 & 35.2 & 35.2 & 35.3 & 35.3 & 35.5 & 35.4 \\
        & en-nl & 27.8 & 28.4 & 28.2 & 28.1 & 27.9 & 28.8 & 28.3 & 28.3 & 28.0 & 27.9 & 28.3 & 28.1 & 27.7 & 27.9 & 28.4 & 28.0 \\
        & nl-en & 31.5 & 30.9 & 31.2 & 31.2 & 31.3 & 30.9 & 31.4 & 31.2 & 30.8 & 30.8 & 30.7 & 30.8 & 31.1 & 31.1 & 30.9 & 31.0 \\
        & en-ru & 29.4 & 29.6 & 29.9 & 29.6 & 30.1 & 29.8 & 30.0 & 30.0 & 29.9 & 30.0 & 29.2 & 29.7 & 30.0 & 29.5 & 29.5 & 29.7 \\
        & ru-en & 35.1 & 34.6 & 35.1 & 34.9 & 34.9 & 34.9 & 35.2 & 35.0 & 34.8 & 34.9 & 35.2 & 35.0 & 34.8 & 34.8 & 35.0 & 34.9 \\
        & en-zh & 41.2 & 40.9 & 40.9 & 41.0 & 41.2 & 40.9 & 40.8 & 41.0 & 40.8 & 40.5 & 40.7 & 40.7 & 40.7 & 40.7 & 41.0 & 40.8 \\
        & zh-en & 46.4 & 46.0 & 46.1 & 46.2 & 46.7 & 46.3 & 46.2 & 46.4 & 46.1 & 46.3 & 46.1 & 46.2 & 46.7 & 46.6 & 46.0 & 46.4 \\
        & avg. & 34.0 & 33.8 & 33.9 & \textbf{33.9} & 34.2 & 34.0 & 34.1 & \textbf{34.1} & 33.9 & 33.8 & 33.8 & \textbf{33.8} & 33.9 & 33.8 & 34.0 & \textbf{33.9} \\
        \bottomrule
    \end{tabular}
    }
    \caption{\textbf{BLEU scores of OPUS in supervised directions}. Scores in \textbf{bold} are the results reported in Table~\ref{tab:bleu1}. ``1,'' ``10,'' and ``20'' indicates three random seeds. ``Res.'' indicates the residual connection of self-attention in the $4^{th}$ encoder layer.}
    \label{tab:bleu5}
\end{table*}

\begin{table*}[t]
    \centering
    \resizebox{\linewidth}{!}{
    \begin{tabular}{l|l|rrrr|rrrr|rrrr|rrrr}
        \toprule
        Layer & \multirow{2}{*}{Direction} & \multicolumn{4}{c|}{S-ENC-T-DEC w/ Res.} & \multicolumn{4}{c|}{T-ENC w/ Res.} & \multicolumn{4}{c|}{S-ENC-T-DEC w/o Res.} & \multicolumn{4}{c}{T-ENC w/o Res.} \\
        Norm & & 1 & 10 & 20 & avg. & 1 & 10 & 20 & avg. & 1 & 10 & 20 & avg. & 1 & 10 & 20 & avg. \\
        \toprule
        \multirow{7}{*}{Pre.} & it-nl & 5.2 & 3.7 & 4.3 & 4.4 & 13.4 & 14.4 & 14.0 & 13.9 & 6.4 & 3.6 & 13.8 & 7.9 & 16.3 & 17.7 & 17.2 & 17.1 \\
        & nl-it & 5.5 & 4.3 & 4.3 & 4.7 & 13.9 & 14.7 & 14.4 & 14.3 & 6.1 & 4.6 & 10.8 & 7.2 & 15.5 & 17.0 & 17.1 & 16.5 \\
        & it-ro & 5.5 & 5.7 & 5.1 & 5.4 & 13.4 & 13.5 & 14.4 & 13.8 & 7.8 & 7.4 & 14.2 & 9.8 & 16.0 & 16.6 & 16.9 & 16.5 \\
        & ro-it & 7.2 & 5.5 & 5.3 & 6.0 & 14.9 & 15.1 & 15.4 & 15.1 & 7.1 & 4.3 & 11.4 & 7.6 & 17.8 & 18.1 & 18.4 & 18.1 \\
        & nl-ro & 4.5 & 4.9 & 4.2 & 4.5 & 12.1 & 12.5 & 12.4 & 12.3 & 6.1 & 7.1 & 11.8 & 8.3 & 12.8 & 14.1 & 14.1 & 13.7 \\
        & ro-nl & 4.4 & 4.3 & 3.9 & 4.2 & 12.1 & 13.4 & 12.5 & 12.7 & 5.6 & 3.1 & 12.4 & 7.0 & 15.1 & 16.1 & 15.6 & 15.6 \\
        & avg. & 5.4 & 4.7 & 4.5 & \textbf{4.9} & 13.3 & 13.9 & 13.9 & \textbf{13.7} & 6.5 & 5.0 & 12.4 & \textbf{8.0} & 15.6 & 16.6 & 16.6 & \textbf{16.2} \\
        \hline
        \multirow{7}{*}{Post.} & it-nl & 13.7 & 11.8 & 13.1 & 12.9 & 15.9 & 16.3 & 17.0 & 16.4 & 17.7 & 18.3 & 17.4 & 17.8 & 18.4 & 18.0 & 18.6 & 18.3 \\
        & nl-it & 14.5 & 12.8 & 12.2 & 13.2 & 15.7 & 17.0 & 16.1 & 16.3 & 18.0 & 18.5 & 18.4 & 18.3 & 17.9 & 18.3 & 18.3 & 18.2 \\
        & it-ro & 12.3 & 11.2 & 12.4 & 12.0 & 14.8 & 14.3 & 15.8 & 15.0 & 17.0 & 17.3 & 17.0 & 17.1 & 17.9 & 17.8 & 18.2 & 18.0 \\
        & ro-it & 14.6 & 13.7 & 13.0 & 13.8 & 17.2 & 16.8 & 17.5 & 17.2 & 19.5 & 20.0 & 20.0 & 19.8 & 19.2 & 19.8 & 20.8 & 19.9 \\
        & nl-ro & 11.1 & 10.4 & 10.2 & 10.6 & 13.5 & 13.4 & 13.6 & 13.5 & 14.9 & 14.9 & 14.7 & 14.8 & 15.4 & 15.2 & 15.5 & 15.4 \\
        & ro-nl & 12.3 & 10.9 & 12.2 & 11.8 & 14.5 & 15.0 & 15.2 & 14.9 & 16.5 & 16.6 & 16.0 & 16.4 & 16.9 & 16.2 & 17.1 & 16.7 \\
        & avg. & 13.1 & 11.8 & 12.2 & \textbf{12.4} & 15.3 & 15.5 & 15.9 & \textbf{15.5} & 17.3 & 17.6 & 17.3 & \textbf{17.4} & 17.6 & 17.6 & 18.1 & \textbf{17.8} \\
        \bottomrule
    \end{tabular}
    }
    \caption{\textbf{BLEU scores of IWSLT in ZST directions}. Scores in \textbf{bold} are the results reported in Table~\ref{tab:bleu1}. ``1,'' ``10,'' and ``20'' indicates three random seeds. ``Res.'' indicates the residual connection of self-attention in the $4^{th}$ encoder layer.}
    \label{tab:bleu6}
\end{table*}

\begin{table*}[t]
    \centering
    \resizebox{\linewidth}{!}{
    \begin{tabular}{l|l|rrrr|rrrr|rrrr|rrrr}
        \toprule
        Layer & \multirow{2}{*}{Direction} & \multicolumn{4}{c|}{S-ENC-T-DEC w/ Res.} & \multicolumn{4}{c|}{T-ENC w/ Res.} & \multicolumn{4}{c|}{S-ENC-T-DEC w/o Res.} & \multicolumn{4}{c}{T-ENC w/o Res.} \\
        Norm & & 1 & 10 & 20 & avg. & 1 & 10 & 20 & avg. & 1 & 10 & 20 & avg. & 1 & 10 & 20 & avg. \\
        \toprule
        \multirow{7}{*}{Pre.} & en-it & 33.9 & 33.8 & 33.6 & 33.8 & 33.7 & 33.4 & 33.7 & 33.6 & 33.6 & 32.9 & 33.3 & 33.3 & 32.4 & 33.3 & 33.4 & 33.0 \\
        & it-en & 37.5 & 37.1 & 37.1 & 37.2 & 37.4 & 37.2 & 37.0 & 37.2 & 35.8 & 36.3 & 36.5 & 36.2 & 35.8 & 36.7 & 36.5 & 36.3 \\
        & en-nl & 29.6 & 29.5 & 29.4 & 29.5 & 29.6 & 29.5 & 29.6 & 29.6 & 29.2 & 29.7 & 29.5 & 29.5 & 29.0 & 29.2 & 29.2 & 29.1 \\
        & nl-en & 31.9 & 32.4 & 32.0 & 32.1 & 32.0 & 32.1 & 31.9 & 32.0 & 30.9 & 31.3 & 31.7 & 31.3 & 31.2 & 31.5 & 31.5 & 31.4 \\
        & en-ro & 24.4 & 25.1 & 25.1 & 24.9 & 25.2 & 25.1 & 25.4 & 25.2 & 24.4 & 24.6 & 24.4 & 24.5 & 24.6 & 24.7 & 24.6 & 24.6 \\
        & ro-en & 31.3 & 31.6 & 31.3 & 31.4 & 32.1 & 31.6 & 31.4 & 31.7 & 30.3 & 30.7 & 30.9 & 30.6 & 30.3 & 31.2 & 31.2 & 30.9 \\
        & avg. & 31.4 & 31.6 & 31.4 & \textbf{31.5} & 31.7 & 31.5 & 31.5 & \textbf{31.6} & 30.7 & 30.9 & 31.1 & \textbf{30.9} & 30.6 & 31.1 & 31.1 & \textbf{30.9} \\
        \hline
        \multirow{7}{*}{Post.} & en-it & 33.9 & 33.3 & 33.5 & 33.6 & 33.8 & 34.0 & 33.5 & 33.8 & 33.1 & 33.2 & 32.6 & 33.0 & 32.4 & 32.6 & 33.4 & 32.8 \\
        & it-en & 37.1 & 36.9 & 37.0 & 37.0 & 37.1 & 37.1 & 36.9 & 37.0 & 35.7 & 35.4 & 36.1 & 35.7 & 36.4 & 35.7 & 35.8 & 36.0 \\
        & en-nl & 29.6 & 30.1 & 30.1 & 29.9 & 30.4 & 30.4 & 30.0 & 30.3 & 29.2 & 29.0 & 29.0 & 29.1 & 29.2 & 29.0 & 29.5 & 29.2 \\
        & nl-en & 31.9 & 32.0 & 31.6 & 31.8 & 31.3 & 31.9 & 31.8 & 31.7 & 31.0 & 31.1 & 31.7 & 31.3 & 30.9 & 30.7 & 31.3 & 31.0 \\
        & en-ro & 25.4 & 25.2 & 24.6 & 25.1 & 25.3 & 25.2 & 25.5 & 25.3 & 24.7 & 25.0 & 24.6 & 24.8 & 24.4 & 24.4 & 25.0 & 24.6 \\
        & ro-en & 31.5 & 31.6 & 31.6 & 31.6 & 30.8 & 31.4 & 31.1 & 31.1 & 30.4 & 29.6 & 30.8 & 30.3 & 30.4 & 30.1 & 30.4 & 30.3 \\
        & avg. & 31.6 & 31.5 & 31.4 & \textbf{31.5} & 31.5 & 31.7 & 31.5 & \textbf{31.5} & 30.7 & 30.6 & 30.8 & \textbf{30.7} & 30.6 & 30.4 & 30.9 & \textbf{30.6} \\
        \bottomrule
    \end{tabular}
    }
    \caption{\textbf{BLEU scores of IWSLT in supervised directions}. Scores in \textbf{bold} are the results reported in Table~\ref{tab:bleu1}. ``1,'' ``10,'' and ``20'' indicates three random seeds. ``Res.'' indicates the residual connection of self-attention in the $4^{th}$ encoder layer.}
    \label{tab:bleu7}
\end{table*}

\begin{table*}[t]
    \centering
    \resizebox{\linewidth}{!}{
    \begin{tabular}{l|l|rrrr|rrrr|rrrr|rrrr}
        \toprule
        Layer & \multirow{2}{*}{Direction} & \multicolumn{4}{c|}{S-ENC-T-DEC w/ Res.} & \multicolumn{4}{c|}{T-ENC w/ Res.} & \multicolumn{4}{c|}{S-ENC-T-DEC w/o Res.} & \multicolumn{4}{c}{T-ENC w/o Res.} \\
        Norm & & 1 & 10 & 20 & avg. & 1 & 10 & 20 & avg. & 1 & 10 & 20 & avg. & 1 & 10 & 20 & avg. \\
        \toprule
        \multirow{13}{*}{Pre.} & es-de & 23.2 & 22.0 & 16.1 & 20.4 & 26.7 & 26.9 & 27.3 & 27.0 & 6.2 & 14.1 & 11.2 & 10.5 & 24.9 & 28.5 & 28.3 & 27.2 \\
        & de-es & 30.3 & 30.0 & 27.6 & 29.3 & 32.4 & 32.0 & 32.3 & 32.2 & 15.5 & 25.7 & 18.7 & 20.0 & 32.9 & 33.1 & 33.4 & 33.1 \\
        & es-fr & 35.0 & 35.6 & 34.0 & 34.9 & 38.8 & 38.8 & 39.3 & 39.0 & 27.8 & 29.8 & 28.2 & 28.6 & 39.9 & 39.8 & 39.9 & 39.9 \\
        & fr-es & 36.0 & 35.5 & 32.8 & 34.8 & 38.6 & 38.7 & 38.7 & 38.7 & 18.7 & 30.7 & 22.3 & 23.9 & 39.7 & 39.7 & 40.0 & 39.8 \\
        & es-nl & 22.7 & 23.0 & 14.2 & 20.0 & 26.4 & 26.3 & 26.3 & 26.3 & 7.0 & 12.8 & 15.0 & 11.6 & 23.2 & 27.7 & 27.5 & 26.1 \\
        & nl-es & 27.2 & 27.1 & 24.9 & 26.4 & 29.1 & 29.1 & 29.1 & 29.1 & 13.9 & 23.0 & 16.9 & 17.9 & 29.6 & 29.7 & 29.8 & 29.7 \\
        & de-fr & 28.6 & 28.1 & 26.9 & 27.9 & 31.4 & 31.3 & 31.7 & 31.5 & 21.9 & 23.0 & 22.5 & 22.5 & 31.9 & 32.3 & 32.2 & 32.1 \\
        & fr-de & 23.5 & 22.0 & 15.9 & 20.5 & 26.3 & 26.5 & 26.8 & 26.5 & 6.3 & 14.3 & 11.5 & 10.7 & 25.0 & 28.1 & 28.2 & 27.1 \\
        & de-nl & 23.2 & 23.4 & 15.0 & 20.5 & 26.3 & 26.2 & 26.0 & 26.2 & 7.0 & 12.8 & 16.2 & 12.0 & 22.5 & 27.5 & 27.2 & 25.7 \\
        & nl-de & 21.4 & 20.3 & 14.3 & 18.7 & 23.2 & 23.8 & 23.5 & 23.5 & 6.4 & 13.3 & 11.9 & 10.5 & 21.6 & 24.6 & 24.6 & 23.6 \\
        & fr-nl & 22.9 & 23.3 & 14.1 & 20.1 & 26.0 & 25.9 & 25.8 & 25.9 & 6.8 & 12.2 & 15.3 & 11.4 & 21.6 & 27.4 & 27.1 & 25.4 \\
        & nl-fr & 26.0 & 25.9 & 25.0 & 25.6 & 28.1 & 28.3 & 28.2 & 28.2 & 19.9 & 20.9 & 19.9 & 20.2 & 28.9 & 28.8 & 28.7 & 28.8 \\
        & avg. & 26.7 & 26.4 & 21.7 & \textbf{24.9} & 29.4 & 29.5 & 29.6 & \textbf{29.5} & 13.1 & 19.4 & 17.5 & \textbf{16.7} & 28.5 & 30.6 & 30.6 & \textbf{29.9} \\
        \hline
        \multirow{13}{*}{Post.} & es-de & 26.0 & 26.9 & 26.8 & 26.6 & 28.2 & 28.4 & 28.7 & 28.4 & 26.1 & 26.3 & 26.1 & 26.2 & 28.7 & 28.7 & 28.7 & 28.7 \\
        & de-es & 32.3 & 32.6 & 32.1 & 32.3 & 33.2 & 33.7 & 33.5 & 33.5 & 32.7 & 31.9 & 32.1 & 32.2 & 33.5 & 33.3 & 33.5 & 33.4 \\
        & es-fr & 37.7 & 38.8 & 37.5 & 38.0 & 40.2 & 40.0 & 40.1 & 40.1 & 37.9 & 37.8 & 37.7 & 37.8 & 40.1 & 39.9 & 40.5 & 40.2 \\
        & fr-es & 37.8 & 38.5 & 38.2 & 38.2 & 40.0 & 39.9 & 40.1 & 40.0 & 38.4 & 37.7 & 38.0 & 38.0 & 39.7 & 39.7 & 40.1 & 39.8 \\
        & es-nl & 25.6 & 26.0 & 26.2 & 25.9 & 27.9 & 27.7 & 27.8 & 27.8 & 26.0 & 25.7 & 25.5 & 25.7 & 27.8 & 28.0 & 27.9 & 27.9 \\
        & nl-es & 29.3 & 29.3 & 29.1 & 29.2 & 29.8 & 30.0 & 29.6 & 29.8 & 29.4 & 29.0 & 29.2 & 29.2 & 29.7 & 29.8 & 29.8 & 29.8 \\
        & de-fr & 30.6 & 31.7 & 30.8 & 31.0 & 32.8 & 32.8 & 33.1 & 32.9 & 31.0 & 30.7 & 30.8 & 30.8 & 32.9 & 32.4 & 33.3 & 32.9 \\
        & fr-de & 25.9 & 26.4 & 26.6 & 26.3 & 27.8 & 28.6 & 28.8 & 28.4 & 26.3 & 26.0 & 25.1 & 25.8 & 28.2 & 28.5 & 28.3 & 28.3 \\
        & de-nl & 25.8 & 26.0 & 25.9 & 25.9 & 27.5 & 27.7 & 27.5 & 27.6 & 25.7 & 25.6 & 25.5 & 25.6 & 27.8 & 27.6 & 27.5 & 27.6 \\
        & nl-de & 23.5 & 23.4 & 23.9 & 23.6 & 24.2 & 24.6 & 24.4 & 24.4 & 23.6 & 23.5 & 23.2 & 23.4 & 24.4 & 24.5 & 24.5 & 24.5 \\
        & fr-nl & 25.3 & 25.8 & 25.6 & 25.6 & 27.4 & 27.4 & 27.3 & 27.4 & 25.5 & 25.5 & 25.3 & 25.4 & 27.8 & 27.6 & 27.5 & 27.6 \\
        & nl-fr & 28.1 & 28.4 & 27.9 & 28.1 & 29.3 & 29.0 & 29.3 & 29.2 & 28.3 & 28.0 & 27.9 & 28.1 & 29.2 & 29.1 & 29.3 & 29.2 \\
        & avg. & 29.0 & 29.5 & 29.2 & \textbf{29.2} & 30.7 & 30.8 & 30.9 & \textbf{30.8} & 29.2 & 29.0 & 28.9 & \textbf{29.0} & 30.8 & 30.8 & 30.9 & \textbf{30.8} \\
        \bottomrule
    \end{tabular}
    }
    \caption{\textbf{BLEU scores of Europarl in ZST directions}. Scores in \textbf{bold} are the results reported in Table~\ref{tab:bleu1}. ``1,'' ``10,'' and ``20'' indicates three random seeds. ``Res.'' indicates the residual connection of self-attention in the $4^{th}$ encoder layer.}
    \label{tab:bleu8}
\end{table*}

\begin{table*}[t]
    \centering
    \resizebox{\linewidth}{!}{
    \begin{tabular}{l|l|rrrr|rrrr|rrrr|rrrr}
        \toprule
        Layer & \multirow{2}{*}{Direction} & \multicolumn{4}{c|}{S-ENC-T-DEC w/ Res.} & \multicolumn{4}{c|}{T-ENC w/ Res.} & \multicolumn{4}{c|}{S-ENC-T-DEC w/o Res.} & \multicolumn{4}{c}{T-ENC w/o Res.} \\
        Norm & & 1 & 10 & 20 & avg. & 1 & 10 & 20 & avg. & 1 & 10 & 20 & avg. & 1 & 10 & 20 & avg. \\
        \toprule
        \multirow{9}{*}{Pre.} & en-de & 28.0 & 28.0 & 28.3 & 28.1 & 28.2 & 28.2 & 28.4 & 28.3 & 28.0 & 28.1 & 28.4 & 28.2 & 28.5 & 28.5 & 28.3 & 28.4 \\
        & de-en & 35.2 & 35.1 & 35.3 & 35.2 & 35.1 & 35.0 & 35.1 & 35.1 & 34.9 & 35.0 & 35.0 & 35.0 & 34.8 & 35.1 & 35.0 & 35.0 \\
        & en-es & 37.6 & 37.4 & 37.4 & 37.5 & 37.5 & 37.4 & 37.7 & 37.5 & 37.5 & 37.5 & 37.4 & 37.5 & 37.5 & 37.5 & 37.3 & 37.4 \\
        & es-en & 39.3 & 38.9 & 39.0 & 39.1 & 39.0 & 39.0 & 38.9 & 39.0 & 38.8 & 39.0 & 39.1 & 39.0 & 38.6 & 39.0 & 38.9 & 38.8 \\
        & en-fr & 36.2 & 36.6 & 36.5 & 36.4 & 36.5 & 36.4 & 36.8 & 36.6 & 36.3 & 36.4 & 36.5 & 36.4 & 36.7 & 36.7 & 36.2 & 36.5 \\
        & fr-en & 38.2 & 38.2 & 38.0 & 38.1 & 38.0 & 38.2 & 38.0 & 38.1 & 38.0 & 37.9 & 38.2 & 38.0 & 37.8 & 38.2 & 38.0 & 38.0 \\
        & en-nl & 28.5 & 28.8 & 28.7 & 28.7 & 28.8 & 28.7 & 28.6 & 28.7 & 28.5 & 28.6 & 28.6 & 28.6 & 28.3 & 28.6 & 28.3 & 28.4 \\
        & nl-en & 31.7 & 31.6 & 31.5 & 31.6 & 31.5 & 31.7 & 31.9 & 31.7 & 31.6 & 31.3 & 31.6 & 31.5 & 31.3 & 31.7 & 31.6 & 31.5 \\
        & avg. & 34.3 & 34.3 & 34.3 & \textbf{34.3} & 34.3 & 34.3 & 34.4 & \textbf{34.4} & 34.2 & 34.2 & 34.4 & \textbf{34.3} & 34.2 & 34.4 & 34.2 & \textbf{34.3} \\
        \hline
        \multirow{9}{*}{Post.} & en-de & 28.4 & 28.4 & 28.7 & 28.5 & 28.6 & 28.7 & 29.0 & 28.8 & 28.5 & 28.2 & 28.4 & 28.4 & 28.7 & 28.5 & 28.3 & 28.5 \\
        & de-en & 35.2 & 35.0 & 35.5 & 35.2 & 34.8 & 35.1 & 34.9 & 34.9 & 35.2 & 35.2 & 35.0 & 35.1 & 35.1 & 35.1 & 34.7 & 35.0 \\
        & en-es & 37.6 & 37.8 & 37.5 & 37.6 & 37.6 & 37.7 & 37.6 & 37.6 & 37.6 & 37.5 & 37.6 & 37.6 & 37.3 & 37.4 & 37.5 & 37.4 \\
        & es-en & 39.4 & 39.0 & 39.0 & 39.1 & 39.0 & 39.3 & 38.8 & 39.0 & 39.2 & 38.9 & 39.1 & 39.1 & 39.0 & 39.1 & 39.1 & 39.1 \\
        & en-fr & 36.8 & 36.8 & 36.4 & 36.7 & 36.8 & 36.7 & 37.0 & 36.8 & 36.6 & 36.5 & 37.1 & 36.7 & 36.9 & 36.8 & 36.7 & 36.8 \\
        & fr-en & 38.3 & 38.2 & 38.4 & 38.3 & 38.2 & 38.2 & 38.4 & 38.3 & 38.2 & 38.1 & 38.2 & 38.2 & 38.1 & 38.3 & 37.9 & 38.1 \\
        & en-nl & 28.8 & 28.8 & 28.6 & 28.7 & 28.7 & 28.7 & 28.9 & 28.8 & 28.6 & 28.6 & 28.9 & 28.7 & 28.7 & 28.7 & 28.5 & 28.6 \\
        & nl-en & 31.5 & 31.6 & 31.7 & 31.6 & 32.1 & 31.7 & 31.7 & 31.8 & 31.7 & 31.9 & 31.5 & 31.7 & 31.7 & 31.4 & 31.4 & 31.5 \\
        & avg. & 34.5 & 34.5 & 34.5 & \textbf{34.5} & 34.5 & 34.5 & 34.5 & \textbf{34.5} & 34.5 & 34.4 & 34.5 & \textbf{34.4} & 34.4 & 34.4 & 34.3 & \textbf{34.4} \\
        \bottomrule
    \end{tabular}
    }
    \caption{\textbf{BLEU scores of Europarl supervised directions}. Scores in \textbf{bold} are the results reported in Table~\ref{tab:bleu1}. ``1,'' ``10,'' and ``20'' indicates three random seeds. ``Res.'' indicates the residual connection of self-attention in the $4^{th}$ encoder layer.}
    \label{tab:bleu9}
\end{table*}

\begin{table*}[t]
    \centering
    \resizebox{\linewidth}{!}{
    \begin{tabular}{l|lll|rrr|rrr }
        \toprule
        \multirow{2}{*}{\#} & \textbf{Layer} & Language & \multirow{2}{*}{Res.} & \multicolumn{3}{c|}{Zero-shot} & \multicolumn{3}{c}{Supervised} \\
         & \textbf{Norm} & Tag & & OPUS & IWSLT & Europarl & OPUS & IWSLT & Europarl \\
        \toprule
        0 & \multicolumn{3}{c|}{\textit{Pivot}} & 55.8 & 64.6 & 73.8 & - & - & - \\
        \hline
        1 & \textbf{PreNorm} & S-ENC-T-DEC & w/ & 35.9 & 34.6 & 66.5 & 63.8 & 70.6 & 74.9 \\
        2 & \textbf{PostNorm} & S-ENC-T-DEC & w/ & \textbf{49.1} & \textbf{51.2} & \textbf{73.0} & 64.1 & 70.6 & 75.0 \\
        \hline
        3 & \textbf{PreNorm} & T-ENC & w/ & 42.5 & 53.0 & 73.0 & 63.7 & 70.6 & 74.9 \\
        4 & \textbf{PostNorm} & T-ENC & w/ & \textbf{43.8} & \textbf{56.0} & \textbf{73.8} & 64.0 & 70.7 & 75.0 \\
        \hline
        5 & \textbf{PreNorm} & S-ENC-T-DEC & w/o & 44.5 & 41.7 & 50.3 & 63.7 & 70.0 & 74.8 \\
        6 & \textbf{PostNorm} & S-ENC-T-DEC & w/o & \textbf{47.6} & \textbf{60.8} & \textbf{72.9} & 64.0 & 69.7 & 74.9 \\
        \hline
        7 & \textbf{PreNorm} & T-ENC & w/o & 42.5 & 57.1 & 72.5 & 63.6 & 69.9 & 74.8 \\
        8 & \textbf{PostNorm} & T-ENC & w/o & \textbf{43.1} & \textbf{60.2} & \textbf{73.8} & 64.0 & 69.7 & 74.9 \\
        \bottomrule
    \end{tabular}
    }
    \caption{\textbf{BLEURT scores}. We report the mean of three seeds and all the translation directions. ``Res.'' indicates the residual connection of self-attention in the $4^{th}$ encoder layer. We mark better scores between PreNorm and PostNorm in \textbf{bold} for ZST.}
    \label{tab:bleurt}
\end{table*}

\end{document}